\documentclass[a4paper,11pt,twoside,openright]{pads-thesis}

\renewcommand{\gTitle}{Predictive
Object-Centric Process Monitoring}

\renewcommand{\gAuthor}{Timo Rohrer}

\renewcommand{\gHeader}{Bachelor's Thesis}

\renewcommand{\gSupervisor}{
    Anahita Farhang Ghahfarokhi \\
    Mohamed Behery \\
}
\renewcommand{\gExaminer}{
    Professor Gerhard Lakemeyer \\
    Professor Wil van der Aalst
}
\renewcommand{\gInstitutionK}{
    Knowledge-Based Systems Group (KBSG) at i5
}
\renewcommand{\gInstitutionP}{
    Process and Data Science (PADS) Chair at i9
}
\renewcommand{\gRegDate}{2020-11-18}
\renewcommand{\gSubDate}{2021-03-18}

\graphicspath{ {figures/} }

\begin{document}
\pagenumbering{roman}

\gTitlePage

\chapter*{Abstract}
The automation and digitalization of business processes has resulted
in large amounts of data captured in information systems, which can aid
businesses in understanding their processes better, improve workflows, or
provide operational support. By making predictions about ongoing processes,
bottlenecks can be identified and resources reallocated, as well as insights
gained into the state of a process instance (case). Traditionally, data is
extracted from systems in the form of an event log with a single identifying
case notion, such as an order id for an Order to Cash (O2C) process.  However,
real processes often have multiple object types, for example, order, item,
and package, so a format that forces the use of a single case notion does
not reflect the underlying relations in the data. The Object-Centric Event
Log (OCEL) format was introduced to correctly capture this information. The
state-of-the-art predictive methods have been tailored to only traditional
event logs. This thesis shows that a prediction method utilizing Generative
Adversarial Networks (GAN), Long Short-Term Memory (LSTM) architectures, and
Sequence to Sequence models (Seq2seq), can be augmented with the rich data
contained in OCEL. Objects in OCEL can have attributes that are useful in
predicting the next event and timestamp, such as a priority class attribute
for an object type package indicating slower or faster processing. In the
metrics of sequence similarity of predicted remaining events and mean absolute
error (MAE) of the timestamp, the approach in this thesis matches or exceeds
previous research, depending on whether selected object attributes are useful
features for the model. Additionally, this thesis provides a web interface
to predict the next sequence of activities from user input.

\tableofcontents

\cleardoublepage

\pagestyle{fancy}
\pagenumbering{arabic}

\chapter*{Acknowledgments}
At first, I would like to express my deep gratitude to my advisers Anahita
Farhang Ghahfarokhi and Mohamed Behery for their continual support throughout
defining the thesis topic, conducting the research, and writing the thesis.
Our regular meetings provided important guidance, excellent feedback, and
interesting discussions.

I would futhermore like to thank Professor Gerhard Lakemeyer and Professor
Wil van der Aalst for accepting this thesis and taking the position of
examiners. I would also like to thank them for enabling this inter-institute
thesis, where I was able to gather compelling insights into both KBSG and PADS.

Lastly I am thankful for my family and friends for their encouragement and
unfailing support.

\chapter{Introduction} \label{chap:intro}
In our increasingly digitized world, data is being collected everywhere.
Data is not only omnipresent in our personal life but also in production
and manufacturing processes, logistics and supply chains, customer
service and internal communication systems. From placing an order
online, that order being processed at a facility, to it being shipped and
delivered, increasingly these activities are all recorded in fine detail
\cite{van_der_aalst_process_2016}. Industry 4.0 and the Internet of Things
have greatly accelerated the need for innovative solutions dealing with the
data generated and recorded \cite{gilchrist_industry_2016}. The Internet
of things, referring to the network of objects being fitted with sensors
and data reporting capabilities, is a tangible representation of the digital
revolution occurring for the last few decades \cite{madakam_internet_2015}. The
tremendous volume, velocity, and variety of data collected poses new challenges
but also presents new opportunities. Businesses can extract value from the
data by making predictions about future activities and thereby increasing
the efficiency of processes \cite{polato_time_2016}.

Processes are comprised of a collection of events which can be extracted
from information systems that record the data. A single event in a
process includes the activity occurring, the timestamp, and additional
resources related to the event. Predictive business process monitoring
aims to learn from past process data to make predictions about ongoing
cases \cite{van_der_aalst_process_2016}. Predictions can be made for what
the next activity will be and when it would occur, or the total time until
completion of the case. It is a branch of process mining which covers a
collection of techniques supporting process discovery, performance analysis,
and conformance checking.  Process mining combines the traditional process
science of process modeling and analysis with data mining and machine learning
methods from data science \cite{van_der_aalst_process_2016}. Driving the
interest for this research discipline is the colossal collection of data in
consumer and business contexts as well as the need to remain competitive by
improving or rethinking processes with that data. There are multiple types
of process mining techniques \cite{van_der_aalst_process_2016}: process
discovery for learning process models from past data, process conformance for
comparing a process to its model, and process enhancement where the model is
improved based on a process. Several case studies have been done using process mining techniques \cite{mann2020connected,uysal2020process} and some tools have been developed in this area \cite{nik2019bipm,ghahfarokhipython}. Additionally, there are techniques that provide
operational support, which includes making predictions about ongoing cases.

\begin{figure}
\centering
\includegraphics[width=\textwidth]{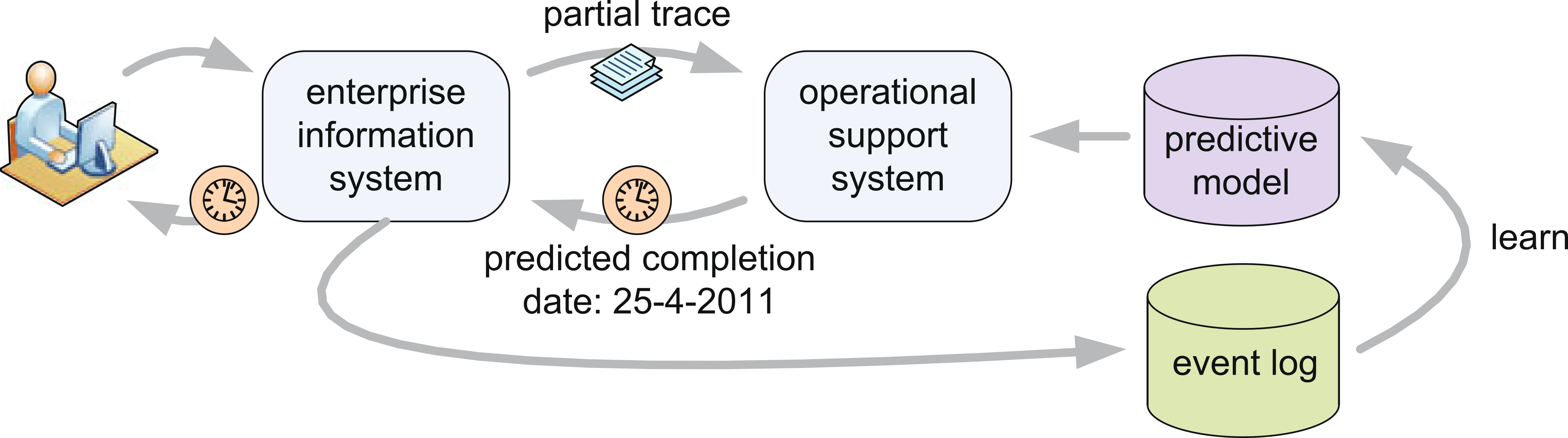}
\caption{Using a model trained on event log data and a partial trace to provide
a prediction about the completion date \cite{van_der_aalst_process_2016}.}
\label{fig:pred}
\end{figure}

By exploiting data of past process instances (cases), predictions can
be made about ongoing ones. As seen in Figure \ref{fig:pred}, based on
event logs a model can, for example, learn to predict the completion time
based on an ongoing case. This thesis focuses on predicting future events,
specifically the activity and the timestamp. However, detecting deviations
in processes as they are occurring or making direct recommendations with
respect to a goal is also possible \cite{van_der_aalst_process_2016}. Being
able to provide such predictions about ongoing cases very powerful. The
time left to resolving customer service cases relaionship management (CRM)
system would be difficult and time-intensive for employees to predict. In
addition to the time to completion, future events themselves can be predicted
\cite{evermann_predicting_2017}. For example, a model might predict that
based on the current ongoing case, an event will occur where additional
information will be requested, or that the case will be resolved shortly.
With automatic predictions, customers requesting that information can easily
be provided with it, with no additional human intervention or work required.
Predicting when an order will be delivered in order to cash (O2C) processes
similarly provides this information to customers without additional effort on
the side of the company. Aside from customer-facing applications, providing
operational support by predicting when production processes will be completed
also provides significant value \cite{folino_discovering_2012}. By giving
estimates about remaining manufacturing timelines based on the current
progress, bottlenecks can be identified and resources re-allocated to improve
efficiency on the fly. By predicting events in a logistics operation, the
scheduling of activities can be optimized to improve the efficiency and
effectiveness of the operation. With the scale and autonomy of business
processes, this form of operational support is vital in the modern
interconnected business world \cite{breuker_comprehensible_2016}.

Processes are often captured by information systems, such as Enterprise
Resource Planning (ERP) and Supply Chain Management (SCM) systems. The stored
data can be extracted as an event log to perform analysis with process mining
tools \cite{van_der_aalst_process_2016}. The event logs that store process
data usually contain at minimum information about the type of activity
occurring, the timestamp of the event, and a case identifier, which refers
to the running case the event is associated with. Traditionally in event
log specifications such as the XES standard, a case notion is required to be
selected for the event log \cite{noauthor_ieee_2016}. The current research
on predictive process monitoring has relied on these traditional event log
standards for training machine learning models \cite{evermann_predicting_2017,
pasquadibisceglie_using_2019, breuker_comprehensible_2016, tax_predictive_2017,
taymouri_predictive_2020, taymouri_encoder-decoder_2020}. However, in many
real information systems capturing process data, there exist multiple objects
that relate to events and a single case notion is inadequate to fully capture
the process and relations occurring \cite{van_der_aalst_object-centric_2019}.

To illustrate the shortcomings of traditional event logs, consider the
following example adapted from \cite{van_der_aalst_object-centric_2019} ;
a customer places an order which is identified with an order id and which
contains multiple different items, each with their own id. The availability
of the different items has to be checked and some items are not yet in
stock. Shortly after, a second order for the same customer is received.
Since all the items of the second order are available, they are packed
together with the available items from the first order into a single package
and delivered. Later, when the remaining items from the first order come into
stock, a second package is loaded and delivered. In total there are multiple
different object types, namely customer, order, item, and package. These
are logged properly by the information system along with any attributes that
exist. But when it comes to extracting the data into an event log, defining
a single identifying case notion is difficult.  Since there are multiple
object types, no one single case notion can capture all the information
contained in this process. \textit{Convergence}, which is when one event
includes different cases. Additionally, in \textit{divergence} problems can arise there  may  be  multiple  instances  of  the same activity within a case. Traditional event
logs make the assumption that there is a single case notion and that each
event refers to only one case. Yet as demonstrated by the example, process
can include multiple object types. More precisely, there exist one-to-many
relations and many-to-many relations between activities and object types
that are not correctly modeled by traditional event logs.

To address this problem object-centric event logs were proposed
\cite{van_der_aalst_object-centric_2019, ghahfarokhi2021ocel}. The shortcomings of traditional
event logs are remedied by not forcing the selection of a single case notion
and preserving one-to-many and many-to-many relations that exist in many
real processes \cite{ghahfarokhi2021process,farhang2022python,berti2022scalable}. The eXtensible Object-Centric (XOC) format was initially
proposed, however, it suffered from complexity and performance problems
\cite{li_extracting_2018}. To address these concerns, the more efficient
OCEL \footnote{http://ocel-standard.org/} standard was specified to aid the
exchange of data between systems and tools. Informally, the data is represented
in two tables, one of the events and one for the objects. The event table is
organized by event id, and holds information about the activity, timestamp,
and other event attributes. Additionally, it includes per event the objects
that are related to the event. So in the example above, the place order
event would include objects of types customer, order, and item
in the OCEL event table. In the object table, the attributes for objects are
recorded. There can be attributes such as the size and weight for an object
of type item, or the address for an object of type customer.

The challenge remains to integrate the OCEL format of event logs with
existing research in prediction methods, as well as using the useful
object attributes in predictions.

\section{Motivation}

The motivation for this thesis is to utilize object-centric event logs
in predicting succeeding events and timestamps in ongoing cases. Since
the related research so far has only used a single case notion format in
event logs, predictive methods have been tailored specifically to that
structure \cite{evermann_predicting_2017, pasquadibisceglie_using_2019,
breuker_comprehensible_2016, tax_predictive_2017, taymouri_predictive_2020,
taymouri_encoder-decoder_2020}. Without selecting a single case notion, the
developed methods can not work. Therefore, in this thesis, the relations
between objects and activities in event logs have to be visualized so
that appropriate case notions can be selected for the desired prediction
task. Additionally, the objects contained in object-centric event logs can
have attributes that aid in predictive tasks. In traditional event logs this
information is a subclass of case attributes \cite{van_der_aalst_process_2016}
and in OCEL logs it is referred to object attributes. For example,
an online order process can contain the object type \textit{package} and
for a given package there can be the object attribute \textit{priority
class}. Such data can be a useful feature for predicting the time until
a package is delivered. While some past research has utilized the case
attributes in traditional event logs \cite{evermann_predicting_2017}, the
more recent state-of-the-art models \cite{taymouri_encoder-decoder_2020}
forgo using this information in predictions.

\section{Research Questions}

The main goal of the thesis is to leverage object-centric event logs to
provide insights for predictions.

To achieve this goal, the questions to be answered are:
\begin{enumerate} 
    \item
        How can recent improvements in machine learning models be enhanced
        with object-centric event logs for prediction?
    \item
        How can training and prediction be made more easily accessible?
\end{enumerate}

\section{Contribution}

The principal contribution of this thesis is bridging the gap between recent
machine learning models and object-centric event logs. Additionally, the
relations between objects and object attributes in OCEL data are used to
improve the prediction. To achieve these goals, this thesis:

\begin{enumerate}
    \item
        Extracts an OCEL dataset from a factory logistics simulation 
    \item
        Augments a state-of-the-art Generative Adversarial Network (GAN)
        approach with object attributes from OCEL data to improve predictions
    \item
        Develops an interactive web tool for making predictions about ongoing
        cases deployed using docker and kubernetes
\end{enumerate}

\section{Thesis Structure}

Chapter \ref{chap:prelim} describes the necessary preliminaries about
process mining, traditional and object-centric event logs, and machine
learning. Chapter \ref{chap:related_work} explores recent research and compares
it to the proposed methods. Chapter \ref{chap:method} details the data driven
prediction model. Chapter \ref{chap:impl} describes the implementation of
the proposed model and the web interface. Chapter \ref{chap:eval} discusses
the results of the evaluation. Chapter \ref{chap:conclusion} summarizes the
work and a future outlook is given.

\chapter{Preliminaries} \label{chap:prelim}
In this chapter, a short introduction of process mining, the necessary
groundwork, and definitions are presented. Formal definitions, as well as
intuitive examples of traditional event logs and object-centric event logs
are given. Lastly, different neural network architectures are explained.

\section{Process Mining}

Processes are ubiquitous in most companies' day-to-day activities. Common
processes include Order to Cash (O2C) and Purchase to Pay (P2P) which
record a businesses' workflow regarding order processing. Furthermore,
manufacturing and production processes are increasingly recorded as factories
move to digitize the assembly line and automate industrial practices
\cite{madakam_internet_2015}. After production, supply chains and logistics
organizations have for long recorded receiving and handling of goods in fine
detail. Beyond that, internal communications, customer service processes,
and essentially any series of activities can and are being recorded for
future analysis.

Recording of the data can occur in a variety of ways: Enterprise Resource
Planning (ERP), Supply Chain Management (SCM), Customer Relationship
Management (CRM), and Business Process Management (BPM) systems record
data about events. Process mining extracts the data in the form of event
logs and aims to extract useful information. Based on the methods and
the final goal, process mining is frequently separated into four types
\cite{van_der_aalst_process_2016}; \textit{Process discovery} deals with
discovering process models from event logs, using algorithms to build
models such as transition systems or Petri nets. \textit{Conformance
checking} deals with detecting abnormalities in an event log as compared
to a process model. \textit{Process reengineering} refers to improving an
existing process model with data from an event log to better reflect the
underlying processes. Lastly, \textit{operational support} is improving
process on the fly by making predictions about remaining time or the
next event and providing recommendations to decrease delays or bottlenecks
\cite{van_der_aalst_time_2011}. Predictive process mining methods fall in the
category of operational support and the developed methods have been tailored
to work with traditional event logs which are described in the next section.

\section{Traditional Event Logs}

Typically the process mining methods assume that information
systems ingest sequential series of events that are recorded with a
variety of information, which can then be extracted as an event log
\cite{van_der_aalst_process_2016}. Table \ref{tab:trad} gives an example
for a traditional event log. Each row in the log shows an event which is
represented by the \textit{event id}. Each row is also uniquely tied to a
case based on a case notion, which is represented by the \textit{case id}. In
addition there is the \textit{activity} and a \textit{timestamp}. While event
logs do not necessarily need to be of this general format, for the purposes
of prediction this information is important.

Since each event is uniquely tied to a case, it can be represented within
a trace. A trace is a finite non-empty sequence of events ordered by time
within a case \cite{van_der_aalst_process_2016}. In Table \ref{tab:trad}
for the case identified by \textit{case id} 73, the activities of the trace
are: \textit{place order, check availability, create package, load package,
create invoice, failed delivery, receive payment, deliver package}.

An event log as it has been traditionally defined is simply a set
of traces, which can also include incomplete or partial traces
\cite{van_der_aalst_process_2016}.

\begin{table}[h]
\caption{A fragment of a traditional event log example (adapted from
\cite{van_der_aalst_object-centric_2019}).}
\label{tab:trad}
\centering
\begin{tabular}{|c|c|c|c|}
\hline
event id & case id & activity & timestamp \\
\hline
\dots & \dots & \dots & \dots \\
9791 & 73 & place order & 2020-9-14 11:37 \\
9792 & 73 & check availability & 2020-9-14 11:40 \\
9793 & 73 & create package & 2020-9-14 15:10 \\
9794 & 74 & place order & 2020-9-14 16:01 \\
9795 & 74 & check availability & 2020-9-14 16:05 \\ 
9796 & 73 & load package & 2020-9-15 10:15 \\
9797 & 73 & create invoice & 2020-9-15 10:16 \\
9798 & 73 & failed delivery & 2020-9-15 14:39 \\
9799 & 74 & create package & 2020-9-15 16:39 \\
9800 & 73 & receive payment & 2020-9-16 09:42 \\
9801 & 75 & place order & 2020-9-17 12:03 \\
9802 & 73 & deliver package & 2020-9-18 10:39 \\
9803 & 75 & check availability & 2020-9-18 11:40 \\
9804 & 76 & place order & 2020-9-18 12:03 \\
9805 & 75 & create package & 2020-9-18 15:10 \\
9806 & 76 & check availability & 2020-9-18 15:40 \\
\dots & \dots & \dots & \dots \\
\hline
\end{tabular}
\end{table}

\section{Object-Centric Event Logs (OCEL)}

The event log shown in Table \ref{tab:trad} can be an inadequate model of
real-world processes. In O2C systems like the examples in Tables \ref{tab:trad}
and \ref{tab:OCELevent}, orders can have multiple items, which might be split
up into multiple different packages with items from other orders. In Table
\ref{tab:OCELevent}, the events that are captured in an object-centric event
log are given. Additionally, there is information about the object types
contained in the events, presented in Table \ref{tab:OCELobj}. The object
types (case notions) are \textit{order, item,} and \textit{package}.

When approaching the data in Table \ref{tab:OCELevent} from
a traditional event log perspective and trying to fit a single
case identifier, events could suffer from convergence or divergence
\cite{van_der_aalst_object-centric_2019}. \textit{Convergence} is when one
event is related to multiple cases and \textit{divergence} occurs when within
a case a group of activities is independently repeated.

Generally, these issues occur at the point of extracting the data from the
systems they were recorded in: different systems such as the SAP ERP software
\footnote{https://www.sap.com/products/erp.html} support multiple object types
and when extracting the data as a traditional event log, the issues arise. If
any one of the object types is picked as the single case identifier, the
problems of convergence and divergence result in an inadequate representation
of the process, since it is not possible to reduce these complex relationships
into a traditional event log \cite{van_der_aalst_object-centric_2019}. By
preserving the complete data in the format of an object-centric event log,
an accurate image of the process is obtained, which can enable higher quality
insights to be obtained with process mining techniques.

\begin{table} 
\caption{A fragment of events in an OCEL example (adapted from
\cite{van_der_aalst_object-centric_2019}).}
\label{tab:OCELevent}
\centering
\begin{tabular}{|c|c|c|c|c|c|}
\hline
event id & activity & timestamp & order & item & package \\
\hline
\dots & \dots & \dots & \dots & \dots & \dots \\
9791 & place order & 2020-9-14 11:37 & $\{o_1\}$ & $\{i_1, i_2, i_3\}$ & $\emptyset$ \\
9792 & check availability & 2020-9-14 11:40 &  $\{o_1\}$ & $\{i_1\}$ & $\emptyset$ \\
9793 & pick item & 2020-9-14 11:41 &  $\{o_1\}$ & $\{i_1\}$ & $\emptyset$ \\
9794 & check availability & 2020-9-14 11:42 &  $\{o_1\}$ & $\{i_2\}$ & $\emptyset$ \\
9795 & check availability & 2020-9-14 11:43 &  $\{o_1\}$ & $\{i_3\}$ & $\emptyset$ \\
9796 & pick item & 2020-9-14 11:47 &  $\{o_1\}$ & $\{i_2\}$ & $\emptyset$ \\
9797 & pick item & 2020-9-14 11:48 &  $\{o_1\}$ & $\{i_3\}$ & $\emptyset$ \\
9798 & create package & 2020-9-14 15:10 & $\emptyset$ & $\{i_1, i_2\}$ & $\{p_1\}$ \\
9799 & place order & 2020-9-14 16:01 & $\{o_2\}$ & $\{i_4,i_5\}$ & $\emptyset$ \\
9800 & check availability & 2020-9-14 16:05 &  $\{o_2\}$ & $\{i_4\}$ & $\emptyset$ \\
9801 & check availability & 2020-9-14 16:06 &  $\{o_2\}$ & $\{i_5\}$ & $\emptyset$ \\
9802 & pick item & 2020-9-15 11:40 &  $\{o_2\}$ & $\{i_1\}$ & $\emptyset$ \\
9803 & pick item & 2020-9-15 11:41 &  $\{o_2\}$ & $\{i_2\}$ & $\emptyset$ \\
9804 & create package & 2020-9-15 16:18 & $\emptyset$ & $\{i_3, i_4, i_5\}$ & $\{p_2\}$ \\
9805 & load package & 2020-9-16 10:15 & $\emptyset$ & $\emptyset$ & $\{p_1\}$ \\
9806 & deliver package & 2020-9-16 14:39 & $\emptyset$ & $\emptyset$ & $\{p_1\}$ \\
9807 & load package & 2020-9-16 16:45 & $\emptyset$ & $\emptyset$ & $\{p_2\}$ \\
9808 & send invoice & 2020-9-17 10:15 & $\{o_1\}$ & $\{i_1, i_2, i_3\}$ & $\emptyset$ \\
9809 & deliver package & 2020-9-17 10:26 & $\emptyset$ & $\emptyset$ & $\{p_2\}$ \\
9810 & send invoice & 2020-9-17 10:45 & $\{o_2\}$ & $\{i_4, i_5\}$ & $\emptyset$ \\
9811 & receive payment & 2020-9-17 10:49 & $\{o_1\}$ & $\{i_1, i_2, i_3\}$ & $\emptyset$ \\
\dots & \dots & \dots & \dots & \dots & \dots \\
\hline
\end{tabular}
\end{table}

\begin{table} 
\caption{A fragment of objects in an OCEL example.}
\label{tab:OCELobj}
\centering
\begin{tabular}{|c|c|c|c|c|c|}
\hline
object id & object type & object priority & object size & object product &
object weight \\
\hline
\dots & \dots & \dots & \dots & \dots & \dots \\
$o_1$ & order & low & NaN & NaN & NaN \\
$o_2$ & order & high & NaN & NaN & NaN \\
$i_1$ & item & NaN & 2mx2mx1m & small box & NaN \\
$i_2$ & item & NaN & 2.4mx2.3mx1.9m & medium box & NaN \\
$i_3$ & item & NaN & 2mx2mx0.5m & small box & NaN \\
$i_4$ & item & NaN & 3mx3mx2m & large box & NaN \\
$p_1$ & package & NaN & NaN & NaN & 50kg \\
\dots & \dots & \dots & \dots & \dots & \dots \\
\hline
\end{tabular}
\end{table}

\newpage

\subsection{OCEL Standard}

To aid in exchanging the data from information systems into formats used in
process mining, the Object-Centric Event Logs (OCEL) standard specification
was introduced \cite{ghahfarokhi_ocel_2020}. The universes used and applicable
examples from Table \ref{tab:OCELevent} and \ref{tab:OCELobj} are given in
Definition \ref{def:univ}.

\begin{definition} [\textbf{Universes}] \label{def:univ}
Below are the universes used:
\begin{itemize}
    \item ${U_e}$ is the universe of event identifiers. \\
        Example: \{9791, 9792, \dots\}
    \item ${U_{act}}$ is the universe of activities. \\
        Example: \{place order, create package, \dots\}
    \item ${U_{att}}$ is the universe of attribute names.
    \item ${U_{val}}$ is the universe of attribute values.
    \item ${U_{typ}}$ is the universe of attribute types.
    \item ${U_{o}}$ is the universe of object identifiers. \\
        Example: \{$i_1, o_1, p_1$, \dots\}
    \item ${U_{ot}}$ is the universe of objects types. \\
        Example: \{package, order, \dots\}
    \item ${U_{timest}}$ is the universe of timestamps. \\
        Example: \{2020-9-16 14:39, \dots \}
\end{itemize}
\end{definition}

Using the universes, an object-centric event log is defined in Definition
\ref{def:OCEL}.

\begin{definition}[\textbf{Object-Centric Event Log}]\label{def:OCEL}
An object-centric event log is a tuple\\ $L = (E, AN, AV, AT, OT, O,
\pi_{typ}, \pi_{act}, \pi_{time}, \pi_{vmap}, \pi_{omap}, \pi_{otyp},
\pi_{ovmap}, \leq)$ such that:

\begin{itemize}
    \item $E \subseteq {U_e}$ is the set of event identifiers.
    \item $AN \subseteq {U_{att}}$ is the set of attributes names.
    \item $AV \subseteq {U_{val}}$ is the set of attribute values (with the
    requirement that $\textrm{AN} \cap \textrm{AV} = \emptyset$).
    \item $AT \subseteq {U_{typ}}$ is the set of attribute types.
    \item $OT \subseteq {U_{ot}}$ is the set of object types. \\
        Example: the object type \textit{item} exists in Table \ref{tab:OCELevent}
    \item $O \subseteq {U_{o}}$ is the set of object identifiers. \\
        Example: the object identifier $p_1$ exists in Table \ref{tab:OCELevent}
    \item $\pi_{typ} : AN \cup AV \rightarrow AT$ is the function associating
    an attribute name or value to its corresponding type.
    \item $\pi_{act} : E \rightarrow {U_{act}}$ is the function associating
    an event (identifier) to its activity. \\
        Example: the first event in Table \ref{tab:OCELevent} had activity \textit{place order}
    \item $\pi_{time} : E \rightarrow {U_{timest}}$ is the function associating
    an event (identifier) to a timestamp. \\
        Example: the first event in Table \ref{tab:OCELevent} had timestamp 2020-9-14 11:37 
    \item $\pi_{vmap}: E \rightarrow (AN \not\rightarrow AV)$ such that
    $$\pi_{typ}(n) = \pi_{typ}(\pi_{vmap}(e)(n)) \quad \forall  e \in E ~
    \forall n \in \textrm{dom}(\pi_{vmap}(e))$$ is the function associating
    an event (identifier) to its attribute value assignments.
    \item $\pi_{omap} : E \rightarrow \mathcal{P}(O)$ is the function
    associating an event (identifier) to a set of related object identifiers. \\
        Example: the first event in Table \ref{tab:OCELevent} is related to
        object types \textit{order, item, package}
    \item $\pi_{otyp} \in O \rightarrow OT$ assigns precisely one object
    type to each object identifier. \\
        Example: the first object in Table \ref{tab:OCELobj} is object type \textit{order}
    \item $\pi_{ovmap} : O \rightarrow (AN \not\rightarrow AV)$ such
    that $$\pi_{typ}(n) = \pi_{typ}(\pi_{ovmap}(o)(n)) \quad \forall n
    \in \textrm{dom}(\pi_{ovmap}(o)) ~ \forall o \in O$$ is the function
    associating an object to its attribute value assignments. \\
        Example: the first object in Table \ref{tab:OCELobj} has attribute value \textit{low}
    \item $\leq$ is a total order (i.e., it respects the antisymmetry,
    transitivity, and connexity properties). A possible way to define a
    total order is to consider the timestamps associated with the events as
    a pre-order (i.e., assuming some arbitrary, but fixed, order for events
    having the same timestamp).
\end{itemize}
\end{definition}

The traditional and object-centric event log standards share a far amount of
similarities but are crucially different in how they treat case notions and
objects. The previous research in predictive methods for event logs discussed
in Chapter \ref{chap:related_work} has only been conducted with traditional
event logs. To better understand the components of the most recent research,
an overview of deep learning is given next.

\newpage

\section{Deep Learning}

Machine learning aims to develop methods to make predictions or decisions after
learning on training data. Regression analysis \cite{van_dongen_cycle_2008},
support vector machines \cite{polato_time_2016}, and extreme gradiant
boosting \cite{senderovich_intra_2017} are models that have performed
well for many different predictive tasks, but the current state-of-the-art
approaches \cite{evermann_predicting_2017, pasquadibisceglie_using_2019,
breuker_comprehensible_2016, tax_predictive_2017, taymouri_predictive_2020,
taymouri_encoder-decoder_2020} utilize neural networks.

\subsection{Neural Networks (NN)}

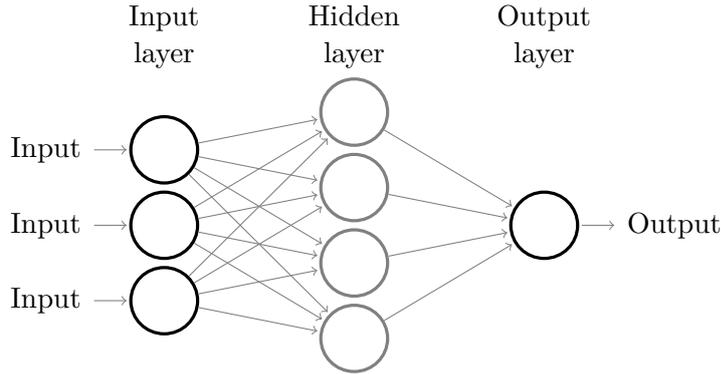
\begin{figure}[ht]
\centering
\begin{tikzpicture}[shorten >=1pt,->,draw=black!50, node distance=2.5cm]
    \tikzstyle{every pin edge}=[<-,shorten <=1pt]
    \tikzstyle{neuron}=[circle,draw=black!50, very thick, minimum size=25pt,inner sep=0pt]
    \tikzstyle{input neuron}=[neuron, draw=black];
    \tikzstyle{output neuron}=[neuron, draw=black];
    \tikzstyle{hidden neuron}=[neuron];
    \tikzstyle{annot} = [text width=4em, text centered]
    \foreach \name / \y in {1,...,3}
        \node[input neuron, pin=left:Input] (I-\name) at (0,-\y) {};
    \foreach \name / \y in {1,...,4}
        \path[yshift=0.5cm]
            node[hidden neuron] (H-\name) at (2.5cm,-\y cm) {};
    \node[output neuron,pin={[pin edge={->}]right:Output}, right of=H-1,yshift=-1.5cm] (O) {};
    \foreach \source in {1,...,3}
        \foreach \dest in {1,...,4}
            \path (I-\source) edge (H-\dest);
    \foreach \source in {1,...,4}
        \path (H-\source) edge (O);
    \node[annot,above of=H-1, node distance=1cm] (hl) {Hidden layer};
    \node[annot,left of=hl] {Input layer};
    \node[annot,right of=hl] {Output layer};
\end{tikzpicture}
\caption{An example of a simple feedforward neural network.}
\label{fig:nn}
\end{figure}

A neural network as seen in Figure \ref{fig:nn} is a network of individual
nodes called artificial neurons \cite{schmidhuber_deep_2015}. Each neuron
receives a number of inputs and produces a single output value.  Neurons are
aggregated into layers and a network usually consists of one input layer, one
output layer, and a number of hidden layers in between. The connections between
nodes are called edges and have associated weights indicating the relative
importance of the connection. The output of an individual neuron is calculated
as the weighted sum of all the inputs and passed through an activation
function, after which it is fed into a number of neurons. The network is
trained on sample data and the weights are adjusted to minimize loss through an
algorithm called backpropagation.

Formally, the activation of a neuron is given in Equation \ref{def:Bact}
\cite{nielsen_neural_2015}:

\begin{equation}
\label{def:Bact}
a^l_j = \sigma \left( \sum_k w^l_{jk} a^{l-1}_k + b^l_j \right)
\end{equation}

Where $a^l_j$ is the activation of the $j$\textsuperscript{th} neuron in the
$l$\textsuperscript{th} layer for a function $\sigma$. The weight matrix $w^l$
consists of the weights connecting to the $l$\textsuperscript{th} layer. The
bias vector $b^l_j$ is the measure of how easy the $j$\textsuperscript{th}
neuron in the $l$\textsuperscript{th} layer fires.

To update the weights and biases a loss or cost function $C$ is defined. It
represents a function used to evaluate the model output. The backpropagation
steps include 4 equations which help calculate the error and gradient of the
loss function $C$. The goal is computing the partial derivatives $\partial C /
\partial w$ and $\partial C / \partial b$. Where $\odot$ is the elementwise
product, the error $\delta^L$ of the output layer $L$ is given in Equation
\ref{def:Berror}:

\begin{equation}
\label{def:Berror}
\delta^L=\nabla_a C \odot \sigma^{\prime}\left(z^L\right)
\end{equation}

Where $\sigma$ is a function and $\nabla_a C$ is a vector comprised of the
components of the partial derivative $\partial C / \partial b^L_j$ and $z^l$
is the weighted input to the neurons in layer $l$. Next the error in terms
of the error in the next layer is given in Equation \ref{def:Berror2}:

\begin{equation}
\label{def:Berror2}
\delta^l = \left( \left( w^{l+1} \right) ^T \delta^{l+1} \right) \odot
\sigma^{\prime} \left( z^{l} \right)
\end{equation}

The element $(w^{l+1})^T$ is the weight matrix transposed. Now the original
goals can be computed; the bias partial derivative as seen in Equation
\ref{def:Bparial}:

\begin{equation}
\label{def:Bparial}
\frac{\partial C}{\partial b^l_j} = \delta^l_j
\end{equation}

And the weight partial derivative give in Equation \ref{def:Bwei}:

\begin{equation}
\label{def:Bwei}
\frac{\partial C}{\partial w} = a^{l-1}_k \delta^l_j
\end{equation}

Backpropagation is presented in Algorithm \ref{algo:back} where
a mini-batch has $m$ samples and the learning rate is $\eta$:

\begin{algorithm}[ht]
    \SetAlgoLined
    \caption{Backpropogation (adapted from \cite{nielsen_neural_2015})}
    \label{algo:back}
    \For{Number of epochs} {
        \For{Number of mini-batches} {
            \ForEach{Sample x in mini-batch} {
                \SetKwFor{ForEach}{foreach}{compute}{end}
                \textbf{compute} $a^{x,1}$\;
                \ForEach{l = 2,3, \dots, L} {
                    $z^{x,l} = w^la^{x,l-1} + b^l$\;
                }
                $a^{x,l} = \sigma(z^{x,l})$\;
                \ForEach{l = L-1, L-2, \dots, 2} {
                    $\delta^{x,L} = \nabla_a C_x \odot \sigma^{\prime}
                    (z^{x,L})$\;
                }
            }
            \SetKwFor{ForEach}{foreach}{update}{end}
            \ForEach{l = L, L-1, \dots, 2} {
                $w^l \rightarrow w^l-\frac{\eta}{m} \sum_{x} \delta^{x, l}
                (a^{x, l-1})^T$\;
                $b^l \rightarrow b^l - \frac{\eta}{m} \sum_x \delta^{x, l}$\;
            }
        }
    }
\end{algorithm}

\subsection{Recurrent Neural Network (RNN)}
\begin{figure}[ht]
\centering
\begin{tikzpicture}[node distance=1.7cm]
  \tikzstyle{box} = [rectangle, minimum width=1cm, minimum height=1cm,text centered, draw=black, fill=gray!20]
  \tikzstyle{equal} = [rectangle, minimum width=1cm, minimum height=1cm,text centered]
  \tikzstyle{blue} = [circle, minimum width=0.5cm, centered, draw=black, fill=black!25]
  \tikzstyle{red} = [circle, minimum width=0.5cm, centered, draw=black, fill=black!25]
  \tikzstyle{arrow} = [thick,->,>=stealth]
  \tikzstyle{dotted} = [dashed, thick,->,>=stealth]

  \node (a1) [box] {R};
  \node (a2) [red, above of = a1] {$h_t$};
  \node (a3) [blue, below of = a1] {$x_t$};

  \node (b4) [box, right of = a1, xshift = 1.5cm] {R};
  \node (b5) [red, above of = b4] {$h_0$};
  \node (b6) [blue, below of = b4] {$x_0$};

  \node (c4) [box, right of = b4] {R};
  \node (c5) [red, above of = c4] {$h_1$};
  \node (c6) [blue, below of = c4] {$x_1$};

  \node (d4) [box, right of = c4] {R};
  \node (d5) [red, above of = d4] {$h_2$};
  \node (d6) [blue, below of = d4] {$x_2$};

  \node (e4) [box, right of = d4, xshift=1cm] {R};
  \node (e5) [red, above of = e4] {$h_t$};
  \node (e6) [blue, below of = e4] {$x_t$};

  \draw [arrow] (a1) -- (a2);
  \draw [arrow] (a3) -- (a1);
  \draw [->, line width=1mm] (a1) -- (b4) node[midway, above] {unfold}; 

  \draw [arrow] (b4) -- (b5);
  \draw [arrow] (b6) -- (b4);
  \draw [arrow] (b4) -- (c4);

  \draw [arrow] (c4) -- (c5);
  \draw [arrow] (c6) -- (c4);
  \draw [arrow] (c4) -- (d4);

  \draw [arrow] (d4) -- (d5);
  \draw [arrow] (d6) -- (d4);
  \draw [dotted] (d4) -- (e4);
 
  \draw [arrow] (e4) -- (e5);
  \draw [arrow] (e6) -- (e4);

  \path[thick,->, >=stealth] (a1) edge [loop left] node {} (a1);

\end{tikzpicture}
\caption{Unfolding a recurrent neural network.}
\label{fig:rnn}
\end{figure}
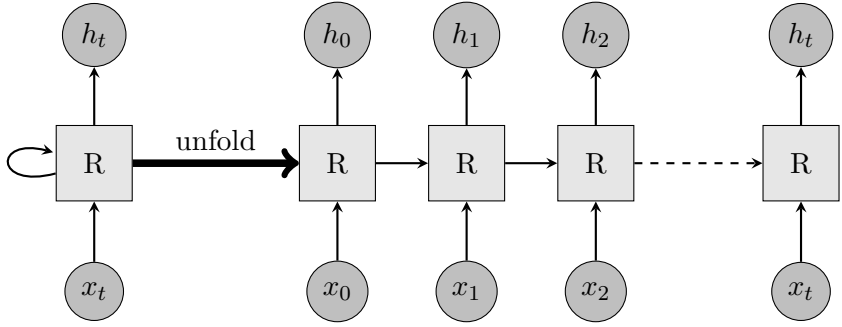

As opposed to feedforward neural networks, where edges do not form any cycles,
there are cyclic structures used in memory like fashion in an RNN which make
them suited well for temporal data \cite{schmidhuber_deep_2015}. Modeled as a
directed graph, they describe a class of networks that can use their internal
state to process inputs of any length. This is achieved by unfolding the RNN
cell, meaning copies are created for different time steps as seen in Figure
\ref{fig:rnn}. The input vector into the RNN $R$ is represented as $x_t$
and the output vector as $h_t$. The output of the cell at t-1 is the input
for the cell at time t. This enables the cell to have memory and process
inputs of different lengths. However, for long sequences, RNNs can suffer
from catastrophic interference where learned patterns are forgotten.

\subsection{Long Short-Term Memory (LSTM)}

\begin{figure}
\centering
\includegraphics[width=0.8\textwidth]{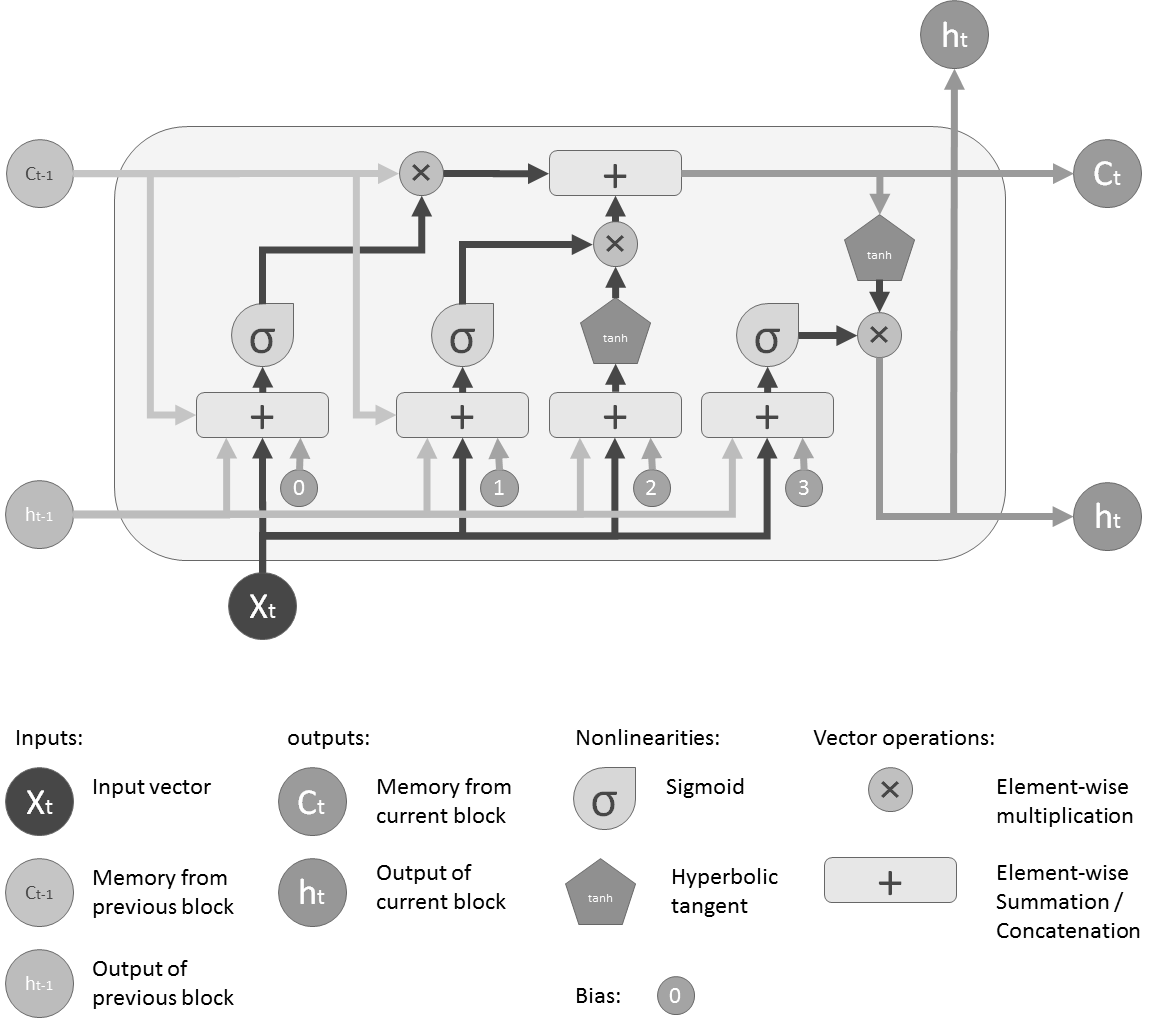}
\caption{LSTM building block \cite{yan_understanding_2017} (Liscensed under GPL-3.0).}
\label{fig:lstm}
\end{figure}

To combat the shortcomings of an RNN, the LSTM architecture was introduced
\cite{hochreiter_long_1997}. It includes gates that control the flow of data
and a memory cell that learns patterns over longer time periods. Intuitively,
the input gate determines how much of the new values flow into the cell,
the forget gate determines how much of the values will be forgotten, while
the output gate controls which values contribute to the output to the next
cell. The gates can be seen as the standard neurons discussed before, which
compute some activation over the sum of weighted inputs.

As seen in Figure \ref{fig:lstm}, the input gate activation vector is $i_t$,
the output gate activation vector $o_t$, and the forget gate activation vector
$f_t$. The activation functions are $\sigma_g$ for the sigmoid function and
$\sigma_h$ for the hyperbolic tangent function. The input vector is $x_t$
and $h_t$ represents the hidden state or output vector. The cell state or
memory vector is represented by $c_t$ and the cell input activation vector
by $\tilde{c}_t$. Finally $W$ and $U$ represent the weight matrices and $b$
any bias that the training step learns. The LSTM cell updates six parameters
in each time step as listed in Equation \ref{def:LSTM}.

\begin{equation}
\label{def:LSTM}
\begin{aligned}
f_t &= \sigma_g(W_{f} x_t + U_{f} h_{t-1} + b_f) &
\tilde{c}_t &= \sigma_h(W_{c} x_t + U_{c} h_{t-1} + b_c) \\
i_t &= \sigma_g(W_{i} x_t + U_{i} h_{t-1} + b_i) &
c_t &= f_t \circ c_{t-1} + i_t \circ \tilde{c}_t \\
o_t &= \sigma_g(W_{o} x_t + U_{o} h_{t-1} + b_o) &
h_t &= o_t \circ \sigma_h(c_t) \\
\end{aligned}
\end{equation}

\subsection{Generative Adversarial Network (GAN)}

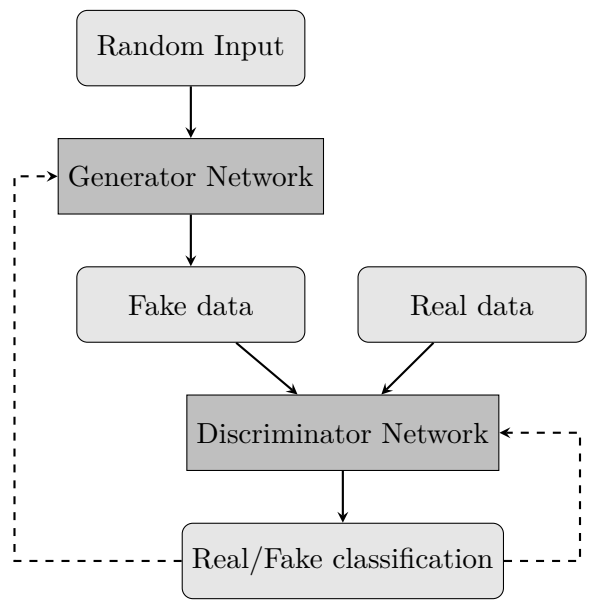
\begin{figure}
\centering
\begin{tikzpicture}[node distance=1.7cm]
  \tikzstyle{box} = [rectangle, rounded corners, minimum width=3cm, minimum height=1cm,text centered, draw=black, fill=gray!20]
  \tikzstyle{disc} = [rectangle, minimum width=3cm, minimum height=1cm,text centered, draw=black, fill=gray!50]
  \tikzstyle{gen} = [rectangle, minimum width=3cm, minimum height=1cm,text centered, draw=black, fill=gray!50]
  \tikzstyle{arrow} = [thick,->,>=stealth]
  \tikzstyle{dotted} = [dashed, thick,->,>=stealth]

  \node (a1) [box] {Random Input};
  \node (a2) [gen, below of = a1] {Generator Network};
  \node (a3) [box, below of = a2] {Fake data};
  \node (a4) [box, right of = a3, xshift=2cm] {Real data};
  \node (a5) [disc, below of = a3, xshift=2cm] {Discriminator Network};
  \node (a6) [box, below of = a5] {Real/Fake classification};

  \draw [arrow] (a1) -- (a2);
  \draw [arrow] (a2) -- (a3);
  \draw [arrow] (a3) -- (a5);
  \draw [arrow] (a4) -- (a5);
  \draw [arrow] (a5) -- (a6);
  \draw [dotted] (a6.east) -- ++ (1cm,0cm) |- (a5.east);
  \draw [dotted] (a6.west) -- ++ (-2.2cm,0cm) |- (a2.west);

\end{tikzpicture}
\caption{Generative adversarial network architecture with random input.}
\label{fig:gan}
\end{figure}

While LSTM's were able to demonstrate good results in predicting the next
event, they require a lot of labeled training data to be able to generalize
well \cite{taymouri_predictive_2020}. Most publically available real-life event
logs are limited in size, which inhibits the model's ability to perform better.
Introduced by Goodfellow et al. in 2014 \cite{goodfellow_generative_2014},
GANs employ a zero-sum game with two player's, meaning one player's gain
is offset by the other player's loss. One player is called the generator
and tries to create new data that is similar to the training data, and the
other player is the discriminator who tries to distinguish between real
and fake data. Each player tries to maximize their outcome through separate
training with backpropagation, which means the generator tries to produce more
convincing fake data and the discriminator tries to get better at detecting
that fake data. Ideally, this converges to the point where the generator
creates near-perfect fakes and the discriminator is no better than 50\% at
predicting the truth. As seen in Figure \ref{fig:gan} the generator is fed with
random data, so early on in the training, the fakes are poor representations
for the real data and the discriminator can easily distinguish them. But
with the feedback from the discriminator, the generator can keep improving.
Radford et al. \cite{radford_unsupervised_2016} standardized deep convolutional
generative adversarial networks (DCGANs) which have been used among other
things, to create realistic human faces and upsample low-resolution images.

\subsection{Sequence to Sequence Model (Seq2seq)}

Sequence to sequence (Seq2seq) models were first introduced in 2014
\cite{sutskever_sequence_2014}. The main problem they address is how to map
one sequence to another. This occurs in many domains, for example, machine
translation, where a sentence in one language will have a different length in
another language. An RNN or an LSTM can not produce variable-length outputs,
so Seq2seq was proposed. There is an encoder/decoder architecture central
to the model: the encoder is an LSTM that turns the input into a fixed hidden
vector and the decoder is another LSTM that transforms the hidden vector
into an output.

\chapter{Related Work} \label{chap:related_work}
In this section, an overview is given for past approaches in predictive
process mining, starting with process model based methods and ending with
the state-of-the-art deep learning models.

\section{Process Model Based Methods}

Process mining is the intersection of data science and process science
\cite{van_der_aalst_process_2016}. Discovering process models to describe past
events has been a topic of research since the 1990's, when to better understand
the execution of business practices an algorithm was developed to generate a
graph model from a log of activities \cite{agrawal_mining_1998}. The proposed
algorithm creates process models and can deal with cycles in processes and
erroneous activities. Notable improvements were made with the $\alpha$-miner
\cite{van_der_aalst_workflow_2004} which can find a Petri net model from an
event log. Furthermore, the heuristic miner \cite{weijters_rediscovering_2003}
can better deal with noisy or incomplete data. This represents process
discovery, where a process model is generated based on event logs. The model
can be visualized to show the workflow between activities.

Apart from performing post-hoc analysis using automatically generated process
models, the use case can be extended to provide real time operational
support of running cases, such as the prediction of the next activity.
The prerequisite for the prediction is a process model obtained from process
discovery methods. Le et al.\cite{le_hybrid_2012} extends higher order Markov
models with a sequence alignment technique to predict the next event. With no
formal description of the underlying processes, the method utilizes probability
matrices for the transition from one event to another. The assumption that
similar sequences are likely to produce the same outcome enables better results
in terms of prediction accuracy. Lakshmanan et al.\cite{lakshmanan_markov_2015}
also first mines a process model from the event log. Then at each split in
the model, a decision tree is learned from execution data and state transition
probabilities computed. Based on an ongoing case a Hidden Markov Model is
used to predict the likelihood of a future event. Parallel executed tasks are
supported and the approach is more accurate as demonstrated on a simulated
auto insurance event log. Breuker et al.\cite{breuker_comprehensible_2016}
recognizes the potential biases in models such as Petri nets and proposes
using probabilistic finite automaton to predict the next event. Based on
research in the field of grammatical interference, a probabilistic model of
the event data is learned and used in next event predictions.

All of these approaches deal with discovering process models from logs based
on the control flow of activities where based on an ongoing case the next
event is predicted. However, with most information systems saving timestamp
data, the models mined from event logs can be enriched with timestamp data.
Then the time until the next event can be predicted, or the remaining time
for the whole case, increasing the usefulness of the predictions.

The time to completion of a case is predicted using non-parametric
regression in \cite{van_dongen_cycle_2008}. Occurrences of activities in
cases, activity durations, and case data are combined in the implementation.
Compared to the naive approach of subtracting the elapsed time from the
total average case time, the proposed methods perform better in real event
logs. The process mining techniques learned from process discovery are
applied to time prediction in \cite{van_der_aalst_time_2011}. A process
model is mined from event logs and augmented with time information. The
annotated transition system predicts completion time based on an ongoing
case, thereby providing operational support. The approach is implemented
in the process mining tool ProM \footnote{https://www.promtools.org/}
and outperforms simple heuristics on both synthetic and real event logs.
In order to support dynamic business processes, ad-hoc predictive clustering
is used in \cite{folino_discovering_2012}. By distinguishing patterns in
event logs as distinct process variants, clusters are discovered. An ongoing
case is estimated to belong to a specific cluster and then the remaining
processing time is predicted. The approach outperforms previous methods on
a real event log and is able to foresee service level agreement violations.
In \cite{rogge-solti_prediction_2013} the authors create Petri net models to
predict remaining time. As opposed to state transition systems, the approach
can model concurrency and time passed is taken into consideration. The
largest gains in predicting accuracy are observed in longer running cases
and it matches previous approaches in shorter cases.

A unifying theme to the approaches presented so far is the reliance on some
form of process model mined through process discovery. Such methods can
not be used when process models are too difficult to obtain. Additionally,
process models can be imperfect abstractions of the underlying process,
trading off precision for simplicity, therefore, making the predictions only
as good as the process model \cite{evermann_predicting_2017}.  Authors in
\cite{verenich_survey_2019} completed an empirical evaluation of remaining
case time prediction methods as of 2019.  They concluded that LSTM Neural
Networks achieved the most accurate results in terms of Mean Average Error,
with the tradeoff of significantly more computational resources being
required to train the models as opposed to transition systems. Recently,
machine learning approaches and specifically deep learning methods have
been developed which outperform previous work for both next event and time
prediction \cite{tax_predictive_2017, taymouri_predictive_2020}. As opposed
to process model based techniques, they do not use an explicit representation
of a process model and instead rely on learning features from the ground up
to make predictions. The large amounts of trainable parameters and inherent
non-linearity in deep neural nets have proven to be advantageous in predictive
tasks \cite{evermann_predicting_2017}.

\section{Deep Learning Based Methods}

Evermann et al.\cite{evermann_predicting_2017} first used deep learning in
process prediction by predicting the next activity. Inspired by research using
Recurrent Neural Networks (RNN) in Natural Language Processing, the approach
is based on LSTM layers and encodes categorical variables. The approach
is tested on multiple real life event logs and is able to surpass previous
research \cite{lakshmanan_markov_2015} \cite{breuker_comprehensible_2016}
in many cases. However, the encoding of attributes in an embedding space
limits the approach to event logs with a small number of unique activities
and numerical values are not utilized. In \cite{pasquadibisceglie_using_2019}
sequences of events are mapped into 2D data structures similar to images
and used to train Convolutional Neural Networks which predict the next
activity. While performing well in terms of precision and recall on multiple
real datasets, less frequent activities are not correctly predicted.

Tax et al. \cite{tax_predictive_2017} combines the challenge of predicting
the next event and the time for that event into one architecture utilizing
LSTMs with two hidden layers. One-hot encoding was used for categorical
data and timestamps were augmented so that business hours were respected
in the prediction. The previous research is surpassed in terms of accuracy
of the predicted next event and Mean Absolute Error (MAE) of the predicted
timestamp on multiple real datasets. Furthermore, by continuously predicting
the next event until the end of a case, this approach can also predict the
entire remaining case and the total time to completion, where it also exceeds
previous approaches. Carmangoe et al. \cite{camargo_learning_2019} combine the
work of \cite{evermann_predicting_2017} and \cite{tax_predictive_2017} to
predict the next event and its timestamp. Categorical attributes are embedded
in an n-dimensional space where coordinates correspond to unique categories
and numerical attributes normalized to reduce variability. Again two LSTM
layers are utilized and different architectures for sharing categorical
attributes are explored.

Taymouri et al. \cite{taymouri_predictive_2020} first introduced the idea
of using Generative Adversarial Nets (GAN) for next event and timestamp
prediction in event logs. By having two LSTM networks play a zero-sum game
against each other the generator becomes better at creating fake next events
over a number of iterations. While normally the generator receives a random
vector from a Gaussian distribution as input, the approach uses the ongoing
case as the input.  The predicted next event from the generator is added on
the ongoing case to form the fake prefix and the real next event is added to
the ongoing case to form the fake prefix. These are fed into the discriminator
which returns a probability of the input being real, which in turn is used
as feedback for the generator. The generative approach allows the network
to make better generalizations about event logs and requires less training
data to produce good results. Activities are one-hot encoded and events are
augmented with the elapsed time between the last event. In both predicting the
next event and predicting the timestamp of the event, the approach was able
to significantly outperform previous approaches on multiple real event logs.

For predicting events further in the future, some of the previously discussed
methods simply execute the prediction consecutively, using the output
from the previous run as input for the current one. While for shorter
average case lengths the results can be good, such methods fare poorly
for longer remaining process executions since the errors propagate. In
\cite{taymouri_encoder-decoder_2020} this is addressed by predicting the
entire remaining events and timestamps from an ongoing case instead of just
the next event. Sequence to Sequence models are integrated into the GAN
architecture from \cite{taymouri_predictive_2020} to map an input prefix
to an output suffix. By utilizing the encoder/decoder structure for the
generator, variable length outputs can be achieved. The generator attempts
to produce convincing suffixes and the discriminator provides feedback by
returning the probability the input is a real suffix for a given prefix.
On several real event logs, the approach performed better in terms of suffix
similarity and error in the predicted timestamps.

\section{Object-Centric Event Logs}

All the presented methods have worked with traditional event logs,
but as \cite{van_der_aalst_object-centric_2019} points out, real
life processes frequently are too complicated to reduce to a single
case notion. Instead, there are multiple identifiers that refer to
different views of the event log and it is not possible to in general
identify a single case identifier. Therefore object-centric event logs
were introduced to more accurately capture the relations in real world
data. Past research in object-centric event logs has been in formalizing
the standard \footnote{http://ocel-standard.org/} and process discovery
\cite{van_der_aalst_object-centric_2019}. Additionally, there have been
techniques proposed in extracting object-centric event logs from information
systems \cite{berti_extracting_2020, simovic_domain-specific_2018} and
automated event log building \cite{de_murillas_case_2020}. Furthermore,
visualizing and modeling the processes has been researched
\cite{van_der_aalst_discovering_2020, cohn_business_2009,
narendra_towards_2009, li_object-centric_2019}. However, the previously
discussed prediction methods rely on traditional event logs with a single case
notion to calculate a trace for a case and train the predictive models. Since
object-centric event logs do not force a single case notion, selecting an
appropriate object type is required first. In this thesis, this is made
easily understandable by showing the relations between activities and object
types as well as statistics about the different object types. Furthermore,
the object attributes contained in OCEL can be useful features in predictive
tasks. While some past research has utilized the analogous case attributes
in traditional event logs \cite{evermann_predicting_2017}, the more recent
state-of-the-art models \cite{taymouri_encoder-decoder_2020} forgo using
this information in predictions.

Therefore, this thesis augments the state-of-the-art GAN with sequence to
sequence models approach \cite{taymouri_encoder-decoder_2020} with object
attributes from object-centric event logs (OCEL). The idea is that certain
object attributes can improve the model, for example, an object attribute
weight for an object of object type package might indicate faster or slower
shipping time. This information can be learned by the model to provide more
accurate predictions. Furthermore, few past research makes the predictive
models easily accessible, therefore, this thesis designs a web interface for
easy training and prediction of OCEL data.

\chapter{GAN for OCEL prediction} \label{chap:method}
In this section, the proposed methods are introduced. First, the necessary
data preprocessing steps and encoding strategy are explained. Second, the
sequence to sequence model architecture for the generator is detailed. Last,
the GAN architecture as a whole is discussed.

\section{Data Preprocessing} \label{datapre}

\begin{figure}[ht]
\centering
\begin{tikzpicture}[node distance=2cm]
    \tikzstyle{activity} = [rectangle, minimum width=3cm, minimum height=1cm,
        text centered, draw=black, fill=blue!30]
    \tikzstyle{type} = [rectangle, minimum width=3cm, minimum height=1cm,
        text centered, draw=black, fill=red!30]
    \tikzstyle{arrow} = [thick]

    \node (place) [activity] {place order};
    \node (check) [activity, below of = place] {check availability};
    \node (pick) [activity, below of = check] {pick item};
    \node (send) [activity, below of = pick] {send invoice};
    \node (receive) [activity, below of = send] {receive payment};
    \node (create) [activity, below of = receive] {create package};
    \node (load) [activity, below of = create] {load package};
    \node (deliver) [activity, below of = load] {deliver package};

    \node (order) [type, right of = check, xshift=7cm, yshift=-1cm] {order};
    \node (items) [type, below of = order, yshift=-2cm] {items};
    \node (package) [type, below of = items, yshift=-2cm] {package};

    \draw [arrow] (place) to node[pos=0.2, above] {1} node[pos=0.9, above] {1} (order);
    \draw [arrow] (place) to node[pos=0.15, above] {1} node[pos=0.84, right] {1..*} (items);
    \draw [arrow] (check) to node[pos=0.1, below] {1..*} node[pos=0.8, above] {1} (order);
    \draw [arrow] (check) to node[pos=0.05, below] {1..*} node[pos=0.8, above] {1} (items);
    \draw [arrow] (pick) to node[pos=0.1, above] {1..*} node[pos=0.8, above] {1} (order);
    \draw [arrow] (pick) to node[pos=0.1, above] {1} node[pos=0.8, above] {1} (items);
    \draw [arrow] (send) to node[pos=0.1, above] {1} node[pos=0.8, above] {1} (order);
    \draw [arrow] (send) to node[pos=0.1, above] {1} node[pos=0.8, above] {1..*} (items);
    \draw [arrow] (receive) to node[pos=0.1, above] {1} node[pos=0.8, above] {1} (order);
    \draw [arrow] (receive) to node[pos=0.1, above] {1} node[pos=0.8, above] {1..*} (items);
    \draw [arrow] (create) to node[pos=0.1, above] {1} node[pos=0.8, above] {1..*} (items);
    \draw [arrow] (create) to node[pos=0.1, above] {1} node[pos=0.8, above] {1} (package);
    \draw [arrow] (load) to node[pos=0.1, above] {1..*} node[pos=0.8, above] {1} (package);
    \draw [arrow] (deliver) to node[pos=0.1, above] {1} node[pos=0.8, above] {1} (package);
    \draw [arrow] (order) to node[pos=0.2, right] {1} node[pos=0.8, right] {1..*} (items);
    \draw [arrow] (items) to node[pos=0.2, right] {1..*} node[pos=0.8, right] {1} (package);

\end{tikzpicture}
\caption{Relations in Table \ref{tab:OCELevent} between activities and object
types (adapted from \cite{van_der_aalst_object-centric_2019}).}
\label{fig:relations}
\end{figure}
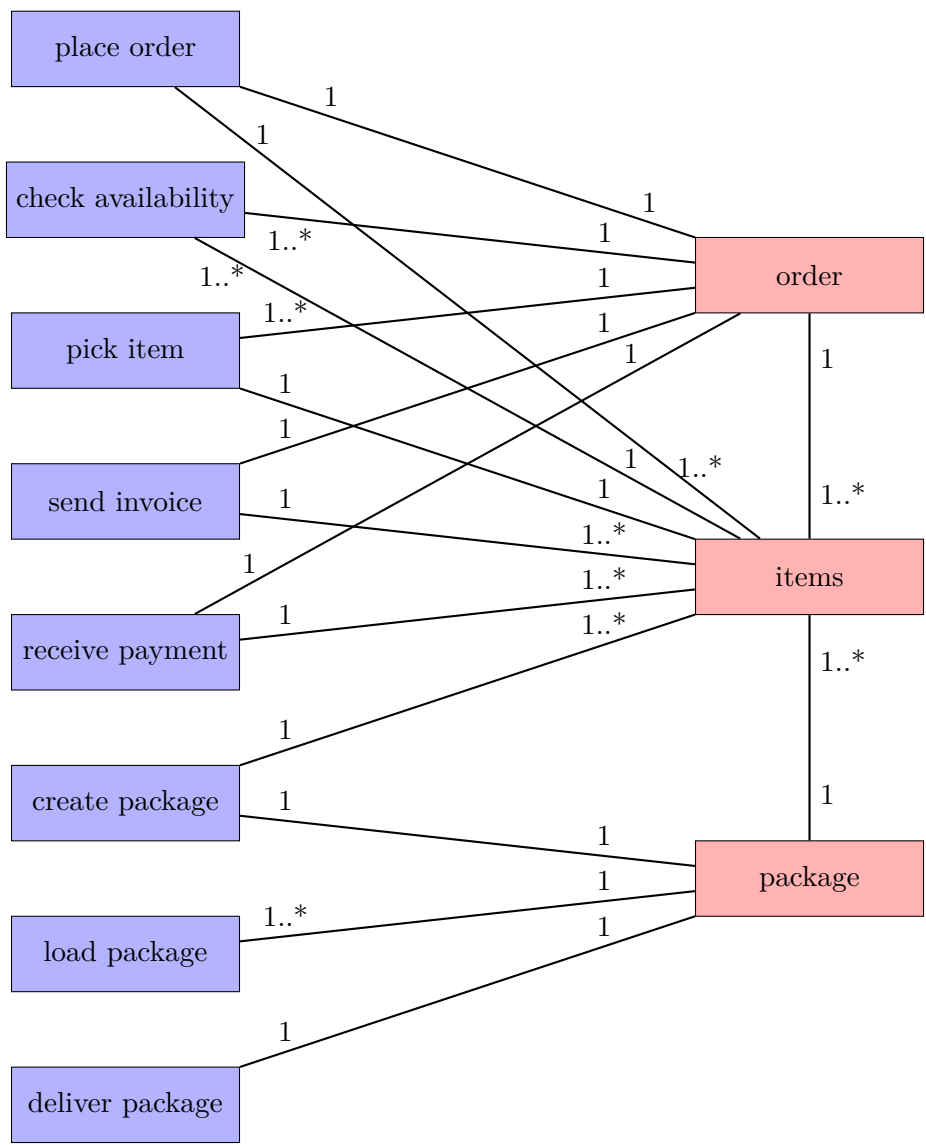

For this thesis, object-centric event logs conforming to the OCEL standard
\cite{ghahfarokhi_ocel_2020} are used. The standard specifies data to be
stored in the XML or JSON format and provides library support for importing
and exporting files in Python. An imported OCEL dataset consists of two
tables, one for the event data (Table \ref{tab:OCELevent}) and one for the
object data (Table \ref{tab:OCELobj}). In the event dataset, object types
have relations with activities. The relations found in the example event
data in Table \ref{tab:OCELevent} are visualized in Figure \ref{fig:relations}.

As seen in Figure \ref{fig:relations}, not every activity is related
to every object type. This will vary from dataset to dataset, so it is
important to visualize the relationships. For example, as seen in Figure
\ref{fig:relations}, only the activities \textit{pack items, load package},
and \textit{deliver package} can be predicted if object type \textit{package}
is selected, since those are the only activities that are related to that
object type. Therefore, depending on the use case of the prediction, the
appropriate object type should be selected. After selection of the object
type \textit{package}, the object Table \ref{tab:OCELobj} contains the object
attribute \textit{object weight} which will be used in the prediction.

The event Table \ref{tab:OCELevent} contains the familiar columns \textit{event
id}, \textit{activity} and \textit{timestamp}. There can
also be additional event attributes. On the other hand, the object Table
\ref{tab:OCELobj}  has the identifier \textit{object id} for the specific
object along with the corresponding \textit{object type}. Furthermore, every
object attribute is listed. For prediction in the model, only specific data
is used: while the \textit{event id} is necessary to organize the cases and
create training or testing data, it is not used directly in the prediction.
Rather, the \textit{activity}, \textit{timestamp}, and any object
attributes are the only values that are used directly in the machine learning
model and therefore need data preprocessing. Two types of data occur in the
OCEL; numerical data such as the \textit{timestamp} and categorical
data such as the \textit{activity}. These require different encoding
so that the model can handle the data.

\subsection*{Numerical Data}

The timestamp of an event occurring is a necessary features for the model,
yet needs conversion to be useful. Instead of utilizing absolute dates and
times, the relative time elapsed between an event and the preceding event
are calculated, as seen in Table \ref{tab:exElapsed} :
\begin{table}[ht!]
\centering
\caption{Example events with calculated elapsed time column.}
\label{tab:exElapsed}
\begin{tabular}{c|c|c}
event id & event timestamp & elapsed (s) \\
\hline
281 & 2019-05-20 09:07:47 & 0 \\
282 & 2019-05-20 09:17:26 & 579 \\
283 & 2019-05-20 11:53:12 & 9346 \\
\end{tabular}
\end{table}

Additionally, the values are normalized as described in Equation \ref{def:norm}
so that the model does not learn false relations about the absolute value
of elapsed time.

\begin{equation}
\label{def:norm}
    \text{normalized value} = \frac{ \text{value} - \text{minimum value}}
    {\text{ maximum value} - \text {minimum value}}
\end{equation}

This data can be encoded in other ways, but to show the approach in this
thesis performs well independent of a special encoding, the simplest one was
chosen.  The \textit{event timestamp} needs to be encoded in this manner,
as well as any other numerical object attributes, such as \textit{object
weight} in Table \ref{tab:OCELobj}.

\subsection*{Categorical Data}

Categorical data such as \textit{event activity} needs to undergo multiple
transformations to become usable in a machine learning model. First, the
data is \textit{label} or \textit{integer} encoded. This refers to converting
values to a number, for example:
\begin{align*}
    \text{place order} &\longrightarrow 1 \\
    \text{confirm order} &\longrightarrow 2 \\
    \text{pay order}  &\longrightarrow  3
\end{align*}
This alone makes categorical data usable in a model, however, the encoding can
be misinterpreted. A model might learn a false hierarchy in the encoding from
the natural order in the numbers. This can lead to the model learning that
2 must come before 1, or because 2 is larger, it must occur more frequently
than 1. These common unintended consequences can be negated by encoding the
numbers obtained using \textit{one hot} encoding.

One hot encoding refers to converting the integers to a representation
similar to a binary encoding, where each category is represented by a column,
as seen in Table \ref{tab:oneHot}:
\begin{table}[ht]
\centering
\caption{Example activites one-hot encoded.}
\label{tab:oneHot}
\begin{tabular}{c|ccc}
event activity & place order & confirm order & pay order \\
\hline
place order & 1 & 0 & 0 \\
confirm order & 0 & 1 & 0 \\
pay order & 0 & 0 & 1 \\
\end{tabular}
\end{table}

The \textit{place order} activity would therefore be encoded as $\langle
1,0,0 \rangle$. The \textit{event activity} needs to be one hot encoded as
well as any categorical object attributes used in the prediction, such as
\textit{object product} in Table \ref{tab:OCELobj}.

\subsection*{Vector Representation}



\begin{definition} [\textbf{Event vector}] \label{def:vector}
    Given an event $e_i \in E$ with the activity $a_i$ where
    $\pi_{act} (e_i)= a_i$ with the timetsamp $t_i$ where $\pi_{time} (e_i)= t_i$, and categorical attributes $C_i \subseteq ATT$ where $\forall c_i \in C_i$ $\pi_{typ} (c_i)= string$ and numerical
    attributes $N_i \subseteq ATT$ where $\forall n_i \in N_i$ $\pi_{typ} (n_i)= float$:
    \begin{itemize}
        \item categorical data $a_i$ and $c_i\in C_i$ is one-hot encoded as $enc(a_i)$ and $enc(c_i)$
        \item elapsed time $l_i$ is calculated from the timestamp $t_i$ and $t_{i-1}$ where $\pi_{time} (e_{i-1})= t_{i-1}$
    \end{itemize}

    The resulting event vector for event $e_i$ is $\langle enc(a_i), enc(c_i), n_i, l_i \rangle$
\end{definition} 

After normalizing the appropriate numerical values and encoding the categorical
values, the event vector from Definition \ref{def:vector} can be built. As
an example, take an event with an activity, both categorical and numerical
object attributes, and a corresponding elapsed time. That example event is
then represented as a vector:
\[
    \text{Event vector: }
    \langle
    \underbrace{0,1,0,0}_\text{activity}
    \underbrace{,0,0,0,1,0,0,0,0,0}_\text{categorical attributes}
    \underbrace{,0.301,1,0.5,0.5818}_\text{numerical attributes}
    \underbrace{,0.04818532}_\text{elapsed}
    \rangle
\]

In previous research \cite{taymouri_encoder-decoder_2020} only the activity
and elapsed time is encoded in the event vector. Object attributes that might
have been present in the underlying process and stored in information systems,
typically are not preserved in traditional event logs, therefore, they can
not be used in the prediction. By leveraging the data contained in OCEL,
this thesis augments the event vector with object attributes, which if the
features are useful in the prediction, can increase the accuracy.

A sequence of events necessitates an end of sequence (EOS) bit, to signal
that the end of the case has been reached. Therefore, the final event is simply
zeros and sets the EOS bit. An example case can be:
\[
    \text{Case: }
    \langle \langle e_1, 0 \rangle, \langle e_2,0 \rangle, \langle e_3,0
    \rangle, \langle 0,1 \rangle \rangle
\]

\subsection*{Prefix/Suffix Split}


\begin{definition} [\textbf{Prefix and Suffix}] \label{def:pref}
    Given a non-empty sequence of $n$ events\\ $\langle e_1,
    e_2, e_3, \dots e_n \rangle$ ordered based on the timestamp and related to an object $o\in O$ where $for~ 1\leq i\leq n$: $o\in \pi_{omap}(e_i)$, a prefix is $\langle e_1, \dots, e_k \rangle$
    for some $k\leq n$. Then a suffix is the remaining events $\langle e_k, \dots,
    e_n \rangle$.
\end{definition} 

When making a prediction, the input is an ongoing process called the
\textit{prefix} and the goal is to predict the remainder of the process called
the \textit{suffix} as defined in Definition \ref{def:pref}. When preparing
the OCEL data, completed processes need to be split up into prefix/suffix
pairs to train the model. For a given process with five events $e_i$ for $1
\leq i \leq 4$, all possible splits are created in Table \ref{tab:split}:

\begin{table}[ht]
\centering
\caption{Making example case into prefix and suffix splits.}
\label{tab:split}
\begin{tabular}{c|c}
prefix & suffix\\
\hline
$e_1$ & $e_2, e_3, e_4$ \\
$e_1, e_2$ & $e_3, e_4$ \\
$e_1, e_2, e_3$ & $e_4$ \\
\end{tabular}
\end{table}

This maximizes the amount of training data available and enables the model
to predict suffixes from both short and long input prefixes.

\section{Generator}

As seen in Section \ref{datapre}, the model needs to be able to take variable
length sequences (prefixes) as an input and produce variable length sequences
(suffixes) as an output. As discussed in Chapter \ref{chap:prelim}, this is
achieved with an encoder/decoder architecture used in Sequence to Sequence
models \cite{sutskever_sequence_2014}, which are the foundation of the
generator. Intuitively the encoder returns a vector based on the input
sequence and the decoder takes that as input and returns the output sequence.

\begin{figure}[ht]
\centering
\begin{tikzpicture}[node distance=2cm]
    \tikzstyle{box} = [rectangle, minimum width=1cm, minimum height=1cm,
        text centered, draw=black, fill=blue!30, rounded corners=0.25cm]
    \tikzstyle{text} = [rectangle, text centered]
    \tikzstyle{arrow} = [thick,->,>=stealth]

    \node (a1) [box] {};
    \node (a2) [box, right of = a1] {};
    \node (a3) [box, right of = a2] {};
    \node (a4) [box, right of = a3] {};

    \node (a5) [box, right of = a4] {};
    \node (a6) [box, right of = a5] {};
    \node (a7) [box, right of = a6] {};

    \node[] (t1) [below of = a1] {$x_1$};
    \node[] (t2) [below of = a2] {$x_2$};
    \node[] (t3) [below of = a3] {$x_3$};
    \node[] (t4) [below of = a4] {$\langle EOS \rangle$};

    \node[] (t5) [above of = a5] {$y_1$};
    \node[] (t6) [above of = a6] {$y_5$};
    \node[] (t7) [above of = a7] {$\langle EOS \rangle$};

    \node[] (l1) [above of = a2] {\textbf{Encoder}};
    \node[] (l2) [below of = a6] {\textbf{Decoder}};

    \draw[arrow] (t1) -- (a1);
    \draw[arrow] (t2) -- (a2);
    \draw[arrow] (t3) -- (a3);
    \draw[arrow] (t4) -- (a4);

    \draw[arrow] (a4) -- (a5) node[midway, above] {$\mathbf{c}_T$};

    \draw[arrow] (a5) -- (t5);
    \draw[arrow] (a6) -- (t6);
    \draw[arrow] (a7) -- (t7);

    \draw[arrow] (a1) -- (a2);
    \draw[arrow] (a2) -- (a3);
    \draw[arrow] (a3) -- (a4);

    \draw[arrow] (a5) -- (a6);
    \draw[arrow] (a6) -- (a7);

\end{tikzpicture}
\caption{Encoder/Decoder seq2seq model mapping $\langle x_1, x_2,
x_3 \rangle$ to $\langle y_1, y_2 \rangle$.}
\label{fig:seq2seq}
\end{figure}
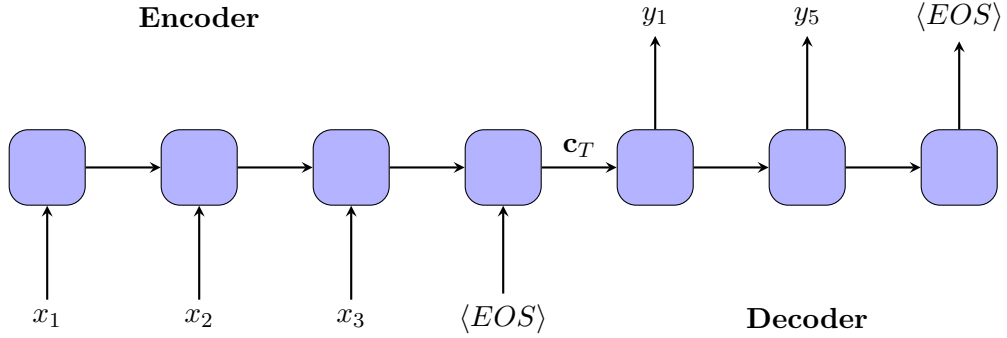

\subsection*{Encoder}
As illustrated in Figure \ref{fig:seq2seq} the input sequence is a vector
containing event vectors and the $\langle EOS \rangle$ to show the end of the sequence. This
is the prefix; the sequence for the ongoing process execution. The encoder is
an LSTM network which processes the prefix input sequence $x_1, \dots, x_t$ of
length $T$. Recall from Definition \ref{def:LSTM} the cell state vector $c_t$:
\[
    c_t = f_t \circ c_{t-1} + i_t \circ \tilde{c}_t 
\]
This is a summary of the input sequence for a timestep. The LSTM cell is
updated $T$ times, meaning the six equations in Definition \ref{def:LSTM}
are calculated, and the final cell state vector $\mathbf{c}_T$ for the
entire prefix is given. As shown in Figure \ref{fig:seq2seq} this
is directly passed into the decoder where it is used
as the initial cell state.

\subsection*{Decoder}
The decoder is another LSTM network that starts generating the output
sequence $y_1, \dots y_{T'}$. Crucially, the output sequence length $T'$
can be different from the input sequence length $T$. In each of the $T'$
decoder updates, the output $y_{t-1}$ from the previous update is used as
the input for the current update.

\subsection*{Conditional Probability}
The goal of the generator therefore, is to estimate the conditional probability
of the output sequence (suffix)  $y_1, \dots, y_{T'}$ given an input sequence
(prefix) $x_1, \dots, y_T$:

This probability given the encoder output $\mathbf{c}_T$ is formally
\cite{sutskever_sequence_2014} given in Equation \ref{def:condProb}:
\begin{equation}
\label{def:condProb}
p(y_1, \dots, y_{T'} \mid x_1, \dots, x_T) = 
\prod_{t=1}^{T'} p(y_t \mid \mathbf{c}_T, y_1, \dots, y_{t-1})
\end{equation}

\newpage

\section{Gumble-Softmax}

GAN architectures require differentiable data in order to update the
generator. However, the vectors that are created from the event logs have
some categorical data in them as well, which when sampled from, are not
differentiable. As demonstrated in \cite{jang_categorical_2017}, a sample
from a categorical distribution can be replaced with a differentiable sample
from a Gumbel-Softmax distribution.

Let $z$ be categorical data with the class probabilities $\pi_1, \pi_2, \dots,
\pi_k$. These are one hot encoded as described before.  With $g_i$ being the
samples from Gumbel(0,1) the Gumbel trick as defined by \cite{maddison__2015}
is given in Equation \ref{def:gumbel}:
\begin{equation}
\label{def:gumbel}
z = \text{one hot} \left( \mathop{arg max}_i [g_i + log (\pi_i)] \right)
\end{equation}

However, the function returns non differentiable values so Softmax is used
to obtain a differentiable approximation to arg max for $y_i$ where $\tau$
refers to the temperature, as seen in Equation \ref{def:gumbelApprox}:
\begin{equation}
\label{def:gumbelApprox}
y_i = \frac{exp((log(\pi_i) + g_i) / \tau)}
{\sum^k_{i=1} exp((log(\pi_j) + g_j) / \tau)}
\end{equation}

\section{GAN Architecture}

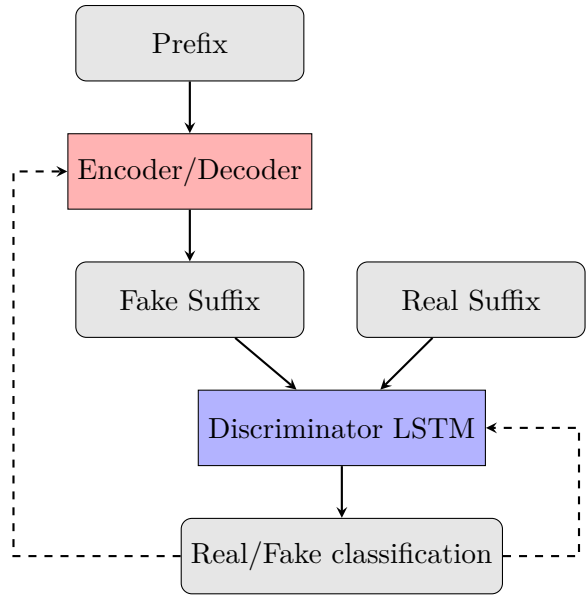
\begin{figure}[ht]
\centering
\begin{tikzpicture}[node distance=1.7cm]
  \tikzstyle{box} = [rectangle, rounded corners, minimum width=3cm, minimum
  height=1cm,text centered, draw=black, fill=gray!20]
  \tikzstyle{disc} = [rectangle, minimum width=3cm, minimum height=1cm,text
  centered, draw=black, fill=blue!30]
  \tikzstyle{gen} = [rectangle, minimum width=3cm, minimum height=1cm,text
  centered, draw=black, fill=red!30]
  \tikzstyle{arrow} = [thick,->,>=stealth]
  \tikzstyle{dotted} = [dashed, thick,->,>=stealth]

  \node (a1) [box] {Prefix};
  \node (a2) [gen, below of = a1] {Encoder/Decoder};
  \node (a3) [box, below of = a2] {Fake Suffix};
  \node (a4) [box, right of = a3, xshift=2cm] {Real Suffix};
  \node (a5) [disc, below of = a3, xshift=2cm] {Discriminator LSTM};
  \node (a6) [box, below of = a5] {Real/Fake classification};

  \draw [arrow] (a1) -- (a2);
  \draw [arrow] (a2) -- (a3);
  \draw [arrow] (a3) -- (a5);
  \draw [arrow] (a4) -- (a5);
  \draw [arrow] (a5) -- (a6);
  \draw [dotted] (a6.east) -- ++ (1cm,0cm) |- (a5.east);
  \draw [dotted] (a6.west) -- ++ (-2.2cm,0cm) |- (a2.west);

\end{tikzpicture}
\caption{GAN architecture for OCEL prediction.}
\label{fig:architecture}
\end{figure}

The generator of the GAN consists of an encoder and decoder and receives a
prefix as input. Conventional GANs use random noise as the input, but since
the goal is to make predictions based on an ongoing process, a prefix is used
instead. The generator returns a fake suffix, which is compared to the real
suffix by the discriminator, as seen in Figure \ref{fig:architecture}. The
discriminator consists of an LSTM and a fully connected layer and returns a
probability representing if the fake suffix resembles a real suffix.

For a prefix $pre$, suffix $suf$, generator $G$, and discriminator
$D$, the objective functions based on \cite{sonderby_amortised_2017}
\cite{taymouri_encoder-decoder_2020} are defined in Definition
\ref{def:objective}:

\begin{definition} [\textbf{Generator and Discriminator objective functions}]
\label{def:objective}
    \[
    \mathcal{L}(D;G) = -log(D(suf)) - log(1-D(G(pre))) \]
    \[
    \mathcal{L}(G;D) = - log \left( \frac{D(G(pre))}{1-D(G(pre))} \right)
    \]
\end{definition}

First, the discriminator $D$ is updated by minimizing the
loss $\mathcal{L}(D;G)$ while the generator stays fixed. When the
discriminator miss-classifies a suffix, the discriminator loss penalizes the
discriminator. This occurs through backprapogation, where the weights of
the network are adjusted through gradient descent.

Next the generator $G$ is updated by minimizing the loss $\mathcal{L}(G;D)$
while the discriminator stays fixed. The generator is penalized
when it produces suffixes that the discriminator is able correctly
classify. Backpropagation occurs through both the discriminator and generator
so the gradients can be calculated, but only the generator's weights are
adjusted in this step.

\chapter{Implementation} \label{chap:impl}
In this section the implementation of Chapter \ref{chap:method} is
detailed for the data preprocessing steps and model initialization and
training. Furthermore, hyperparameter tuning of the model is discussed,
the web interface is described and the deployment is explained.

\section{Extracting RCLL dataset}

The RCLL is a simulation of factory logistics discussed in more detail in
Chapter \ref{chap:eval}. The events that occur are saved in a database with
some unecessary for this thesis information about the route, distance, and
more. The relevant event information of indentifier, activity, and timestamp is
extracted. The object types of robot, order, and components are extracted as
well and the event table of the OCEL standard can be created. The attributes
for the objects are extracted to create the object table. The final dataset
is stored according to the OCEL standard in the JSON file format.

\section{Data preprocessing}

The data preprocessing module handles the steps discussed in Chapter
\ref{chap:method} as well as some common additional functions. The
module was written in Python 3, using the pandas data analysis library
\footnote{https://pandas.pydata.org/} and PyTorch machine learning library
\footnote{https://pytorch.org/}.

\textbf{Importing OCEL} \\
The pm4py-mdl python package \footnote{https://github.com/Javert899/pm4py-mdl}
is utilized to import the OCEL data. Reading both XML and JSON files as per
the OCEL standard is supported and the import methods return two pandas
dataframes. One dataframe represents the events and is similar to Table
\ref{tab:OCELevent} and the other represents the objects and is similar in
form to Table \ref{tab:OCELobj}.

\textbf{Elapsed Time Calculation} \\
The elapsed time is calculated for individual cases in the events dataframe.
The elapsed time for the first event in a case, therefore, is zero and for
every subsequent event, it is the time that passed after the event right before.

\textbf{One-hot Encoding} \\
The categorical values including activity names and categorical object
attributes are one-hot encoded and the mapping is saved for later
reversal. This mapping is also important when predicting for a new prefix,
there the categorical values are encoded with this mapping so the model can
interpret the data correctly.

\textbf{Normalizing} \\
Numerical values used in the prediction are normalized and the terms of
the normalization saved. When using the model in prediction these terms are
needed to reverse the normalization and interpret the results in absolute
timestamp terms.

\textbf{Partitioning} \\
When a mini-batch has randomly chosen inputs, it can have a large amount
of variability in the input lengths. While the model can be updated
more frequently using Mini-Batch Gradient Descent as shown in Algorithm
\ref{algo:back}, the downside is that different length cases cause slowdowns in
the computations, as observed in \cite{sutskever_sequence_2014}.  Therefore,
the partitions are groups of cases with a similar length, which can be
batched together.

\textbf{Variable Length Splits} \\
Within the created partitions, traces are split into every possible
combination of lengths for the prefix and suffix. The prefixes and
corresponding suffixes are saved as PyTorch tensors. The tensors are saved as
TensorDatasets in DataLoader objects. This enables easy access during training.

\section{Model}

The generator consists of two parts; the encoder and the decoder. The encoder
is an LSTM network which maps the inputs to the hidden vector. The decoder
is also an LSTM network connected to a fully connected layer and then passed
through ReLU and Sigmoid activation functions. A concept called teacher
forcing \cite{williams_learning_1989} is used in the generator every so often
based on randomness. When it is triggered, instead of using the output from
the decoder from the previous time step as the input for the current one,
the real suffix is used as the input. This can lead to faster training and
better results. The discriminator is an LSTM network connected to a fully
connected layer.

\subsection{Training}

Training occurs based on the Algorithm \ref{algo:gan}.

\begin{algorithm}[ht]
    \SetAlgoLined
    \caption{Training GAN}
    \label{algo:gan}
    \For{Number of epochs} {
        \For{Number of mini-batches} {
            \For{prefix, suffix in training data} {
                forward pass of generator and discriminator\;
                zero discriminator gradients\;
                update the discriminator to minimize loss\;
                clip discriminator gradient norm\;
                zero generator gradients\;
                update the generator to minimize loss\;
                clip generator gradient norm\;
            }
            \If{epoch multiple of 5} {
                run model on validation data\;
                save model if beats current best\;
            }
        }
    }
\end{algorithm}

The gradients need to be set to zero in lines 3 and 7 because PyTorch
accumulates the gradients on the backward passes. Exploding gradients can
occur when large updates to weights cause numerical under or overflow,
which happens in LSTM networks since they need to unroll many timesteps as
discussed in Chapter \ref{chap:prelim}.  The solution to that is gradient
norm clipping seen in lines 6 and 9.

\subsection{Validation and Testing}

In the data preprocessing step, the main OCEL data is split into training,
validation, and testing sets. Training occurs with the training data, which
represents 70\% of the total data. As seen in Algorithm \ref{algo:gan}, during
training the model is evaluated on validation data periodically. Validation
data represents 10\% of the total data. Last, there is the testing data with
20\% of the total data. The model is benchmarked using either validation
data or testing data in the evaluation function. The model is set to
evaluation mode to prevent weights from being updated and a forward pass
of the model (generator) is performed. The fake suffix returned from the
generator is compared to the real suffix in two metrics, which are discussed
in Chapter \ref{chap:eval}.

\section{Hyperparameter Tuning}

\begin{figure}[ht]
\centering
\includegraphics[width=0.55\textwidth]{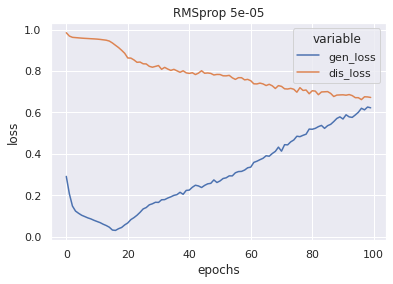}
\caption{Convergence of generator and discriminator loss on synthetic dataset.}
\label{fig:conv}
\end{figure}

Convergence is difficult for GANs due to the nature of the adversarial
game; gains by the generator result in losses by the discriminator and vice
versa. Ideally, the generator improves to the point where the discriminator
has a $50\%$ accuracy. Then, the loss calculated for each should be about
the same, as seen in the convergence graph in Figure \ref{fig:conv}. This
can be problematic since the discriminator continues giving feedback to
the generator and the resulting fake suffixes can decrease in quality. As
a result, the convergence might be shortlived. Training can fail because of
model collapse, where the generator can only output a small number of different
suffixes. Convergence failure is also possible, where the generator loss keeps
increasing while the discriminator loss is near zero, meaning the generator
produces very poor quality suffixes that the discriminator can easily identify.

The model has multiple hyperparameters: learning rate, type of optimizer, and
the number of layers. The optuna python library\footnote{https://optuna.org}
was used to aid in empirically finding the best hyperparameters. The
hyperparameters are suggested by optuna based on a given range and then
multiple trials are conducted to find the best settings. Optuna searches
the range of values efficiently using the tree-structured Parzen Estimator
algorithm and can prune trials that are performing poorly to avoid
wasting resources. The algorithm repeatably decreases the search space,
first searching in the range suggested. Based on the evaluation of the
model with those hyperparameters, the range is narrowed, leading to an
optimal search space. The best hyperparameters found through tuning were:
the RMSprop Algorithm for the optimizer, a learning rate of $5.5e^{-5}$,
and five layers for the LSTM networks. The values were similar to the options
used in past research \cite{taymouri_encoder-decoder_2020}.

\section{Interface}

\begin{figure}[ht]
\centering
\includegraphics[width=0.9\textwidth]{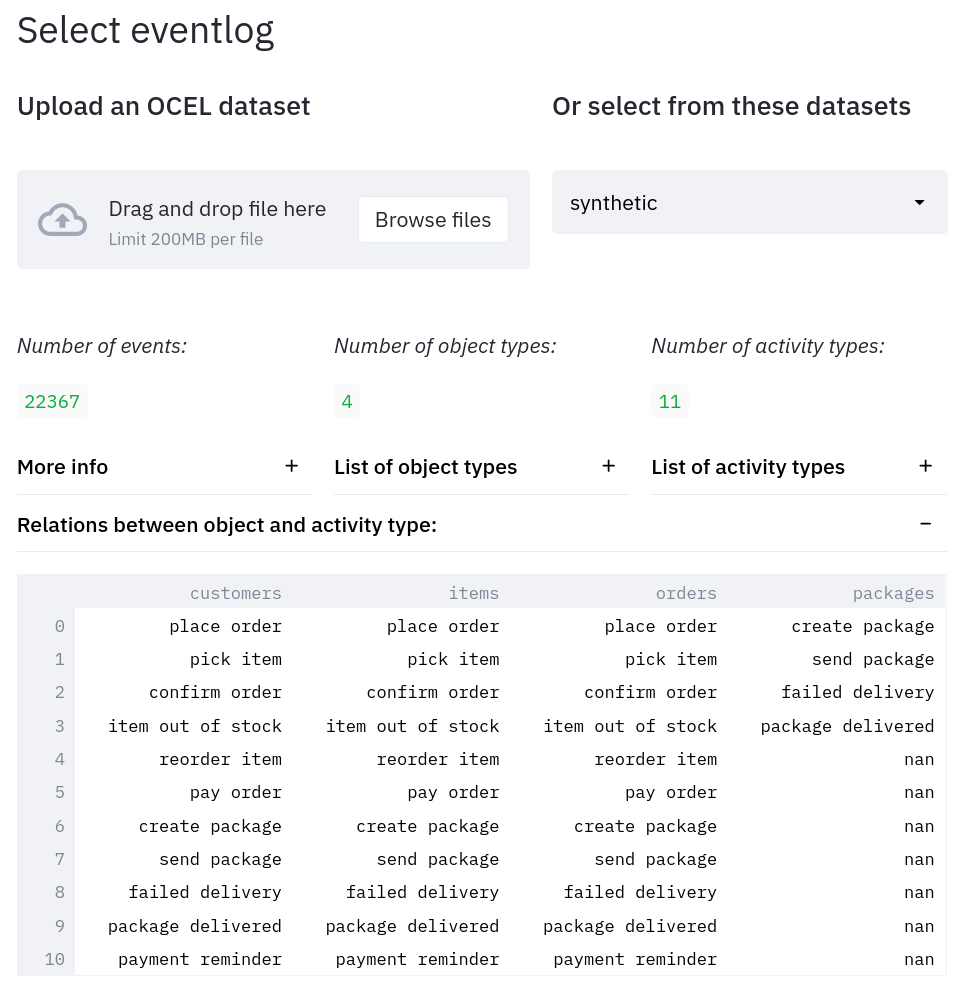}
\caption{Interface section for selecting the event log and viewing information.}
\label{fig:int1}
\end{figure}

The web interface is implemented in Python using the streamlit library
\footnote{https://www.streamlit.io/}. As seen in Figure \ref{fig:int1}, the
first main section of the interface is \textit{Select eventlog}. There either
an event log in the OCEL format can be uploaded, or one of the preloaded
datasets picked from the dropdown menu.  Next, there is an overview of common
statistics regarding the activities and object types present in the data. The
choice of an object type to use in the prediction is vital, therefore, a table
visualizing the relations between activities and object types is presented.
The model can then be trained based on the selected object type and next
the prediction starts.

\begin{figure}[ht]
\centering
\includegraphics[width=0.7\textwidth]{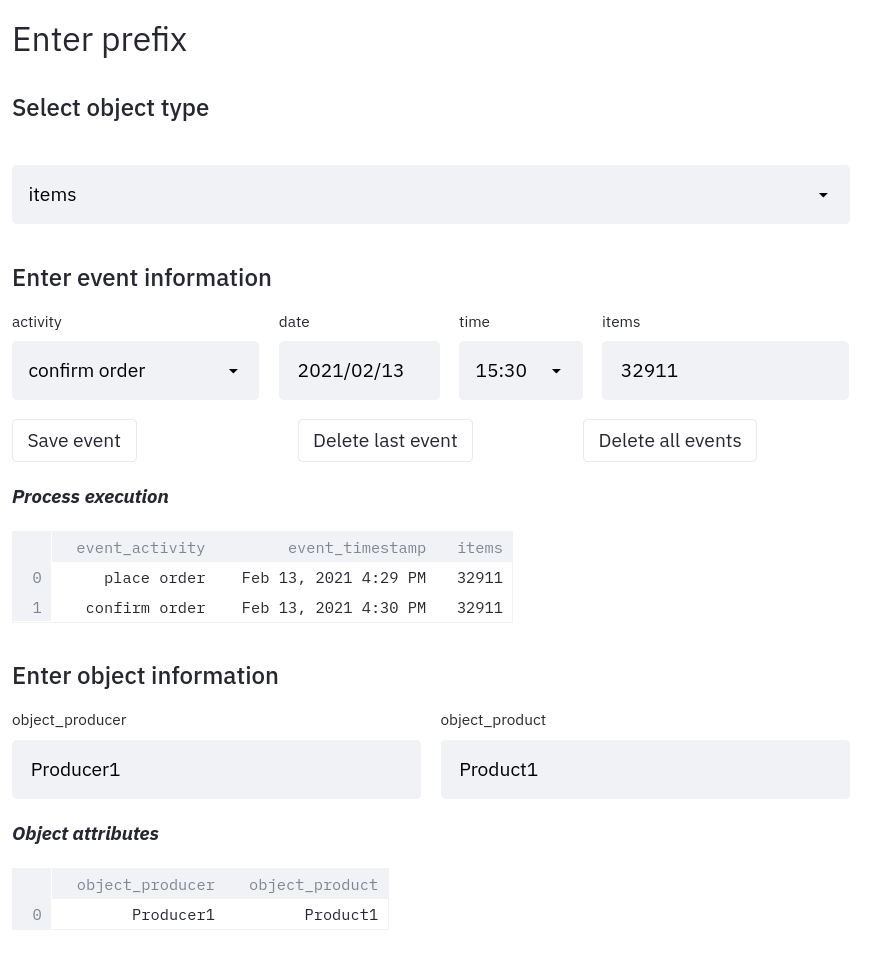}
\caption{Interface section for inputting prefix.}
\label{fig:int2}
\end{figure}

As seen in Figure \ref{fig:int2}, the section for entering a prefix (ongoing
case) starts with selecting the object type. Based on that information, the
dropdown menu \textit{activity} under "Enter event information" is populated
by all the activities that the object type has a relation to. After selecting
the activity, the timestamp information and identifier for the object can
be entered. By choosing the "Save Event" button the event can be saved and
is presented in the table under "Process execution". From there, more events
can be created and saved to create a prefix for which the prediction should
be performed on. Under the section "Enter object information" the fields are
automatically retrieved from the OCEL log, presenting only the object attribute
classes that are related to the object type. There, the desired attribute values
can be entered. That concludes the creation of a prefix with both event and
object attributes. Subsequently, the interface presents the predicted suffix
for the prefix that was entered, using the trained model.

\section{Deployment}

To deploy the model and the interface easily, the container-orchestration
system Kubernetes \footnote{https://kubernetes.io/} is used. The interface
setup and required dependencies are defined in a Dockerfile and a Docker
image \footnote{https://www.docker.com/} is built. This represents the
information required to configure a container in Kubernetes using the
Deployment object. The Kubernetes cluster can then automatically create
replica pods based on current load requirements and the interface is exposed
by the Service object.

\chapter{Evaluation} \label{chap:eval}
In this chapter the the experimental setup is detailed. Then the datasets
used in evaluating the model are presented and the reslts are discussed.

\section{Experimental Setup}

The model was implemented as described in Chapter \ref{chap:impl}.
The training and testing was performed on an Intel Xeon CPU, 16 GB RAM,
and a Nvidia Tesla T4 GPU with 16 GB of GPU memory.

\subsection*{Metrics}

The evaluation metrics are kept the same as previous research
for uniformity. To compare the predicted suffixes to the
real suffixes, the approach for calculating similarity $S$ from
\cite{taymouri_encoder-decoder_2020} is used as described in Equation
\ref{def:seqSim}:

\begin{equation}
\label{def:seqSim}
S(s_f, s_r) = 1 - \left( \frac{DL(s_f, s_r)}{Max(|s_f|, |s_r|)} \right)
\end{equation}

Where $s_f$ is the predicted suffix and $s_r$ is the real suffix.  Since the
suffix is a sequence, Damerau–Levenshtein (DL) distance can be used. It
is defined as the minimum number of operations needed to transform one
sequence into the other sequence. The allowed operations are insertion,
deletion, substitution, or transposition.

As the measure of accuracy of timestamps, Mean Absolute Error is used
(MAE) as seen in Equation \ref{def:mae}:

\begin{equation}
\label{def:mae}
MAE = \frac{\sum^n_{i=1} |t_r - t_f|}{n}
\end{equation}

Where $t_r$ is the real timestamp and $t_f$ the predicted timestamp.

\section{Synthetic Dataset}

To validate the approach a synthetic dataset was used. The synthetic dataset
represents an order process with different object types. Orders with multiple
items are placed by customers. Items are packed into packages and sent
out. An excerpt of the data can be seen in Figure \ref{fig:syntheticFull}
in the Appendix.

\subsection{Description}

The activities modeled in the dataset are:
\begin{center}
\textit{\{place order, pick item, confirm order, item out of stock, reorder
item, pay order, create package, send package, failed delivery,
package delivered, payment reminder\}}
\end{center}

The object types in the dataset are:
\begin{center}
\textit{\{customers, items, orders, packages\}}
\end{center}

The attributes for each object type are shown in Table \ref{tab:synObjTypes}:
\begin{table}[ht]
    \centering
    \caption{Object attributes in synthetic object centric event log.}
    \label{tab:synObjTypes}
    \begin{tabular}{cc} \toprule
    object type & object attribute \\ \midrule
    \textit{customers} & age \\
    \textit{items} & color \\
    \textit{orders} & price \\
    \textit{packages} & weight \\ \bottomrule
    \end{tabular}
\end{table}

An important step in selecting an appropriate object type is understanding
the relations between activities and object types as seen in Table
\ref{tab:synRelations}.

\begin{table}[ht]
    \caption{Relations between activities and object types in the synthetic dataset.}
    \label{tab:synRelations}
    \centering
    \begin{tabular}{llll} \toprule
    \textit{customers}         & \textit{items}             & \textit{orders}            & \textit{packages} \\ \midrule
    \textit{place order}       & \textit{place order}       & \textit{place order}       & $\emptyset$               \\
    \textit{pick item}         & \textit{pick item}         & \textit{pick item}         & $\emptyset$               \\
    \textit{confirm order}     & \textit{confirm order}     & \textit{confirm order}     & $\emptyset$               \\
    \textit{item out of stock} & \textit{item out of stock} & \textit{item out of stock} & $\emptyset$               \\
    \textit{reorder item}      & \textit{reorder item}      & \textit{reorder item}      & $\emptyset$               \\
    \textit{pay order}         & \textit{pay order}         & \textit{pay order}         & $\emptyset$               \\
    \textit{create package}    & \textit{create package}    & \textit{create package}    & \textit{create package}    \\
    \textit{send package}      & \textit{send package}      & \textit{send package}      & \textit{send package}      \\
    \textit{failed delivery}   & \textit{failed delivery}   & \textit{failed delivery}   & \textit{failed delivery}   \\
    \textit{package delivered} & \textit{package delivered} & \textit{package delivered} & \textit{package delivered} \\
    \textit{payment reminder}  & \textit{payment reminder}  & \textit{payment reminder}  & $\emptyset$               \\ \bottomrule
    \end{tabular}
\end{table}

Additionally, to better understand the cases that are captured by the
different object types, some statistics are given in Table \ref{tab:synStats}.
These represent data after outliers outside of two standard deviations in terms
of case length had been removed, which is also what is used when training.
For example, in the case of the object type \textit{customers}, this does
not remove any data, but in the case of \textit{orders}, the number of cases
decreases by less than 1\% and the max length for a case is decreased from
41 to 28.

\begin{table}[ht]
    \caption{Statistics about synthetic object centric event log.}
    \label{tab:synStats}
    \centering
    \begin{tabular}{lllll|lll}\toprule
    &  & \multicolumn{3}{c}{Length of Case} & \multicolumn{3}{c}{Case time (days)} \\
    \cmidrule(lr){3-5}\cmidrule(lr){6-8}
    & Number of cases & Max & Min & Mean & Max & Min & Mean \\ \midrule
    \textit{customers} & 17   & 1470 & 1171 & 1315.7 & 461.03 & 383.7 & 422.5 \\ 
    \textit{items}     & 7860 & 10   & 7    & 7.82   & 103.87 & 0.51  & 14.48 \\ 
    \textit{orders}    & 1923 & 28   & 7    & 15.57  & 114.69 & 1.67  & 18.9  \\ 
    \textit{packages}  & 1238 & 4    & 3    & 3.14   & 7.71   & 0.03  & 1.69  \\ \bottomrule
    \end{tabular}
\end{table}

\subsection{Results}

\begin{table}[ht]
    \caption{Sequence similarity and MAE results for synthetic object centric
    event log.}
    \label{tab:resSyn}
    \centering
    \begin{tabular}{lccc|ccc}\toprule
    & \multicolumn{3}{c}{Sequence similariy $S$} & \multicolumn{3}{c}{MAE (normalized)} \\
    \cmidrule(lr){2-4}\cmidrule(lr){5-7}
              & Thesis  & Taymouri \cite{taymouri_encoder-decoder_2020}
              & Tax\cite {tax_predictive_2017} & Thesis &
              Taymouri\cite {taymouri_encoder-decoder_2020} &
              Tax\cite {tax_predictive_2017} \\ \midrule
    \textit{customers} & 0.0986  & 0.0921   & 0.0371 & 0.4559 & 0.4983 & 0.5839 \\
    \textit{items}     & 0.5102  & 0.5113   & 0.4781 & 0.0384 & 0.0461 & 0.0438 \\
    \textit{orders}    & 0.3502  & 0.3288   & 0.3013 & 0.0336 & 0.0310 & 0.0487 \\
    \textit{packages}  & 0.8654  & 0.8241   & 0.8121 & 0.0673 & 0.0632 & 0.0729 \\ \bottomrule
    \end{tabular}
\end{table}

The predominant trend that can be observed in Table \ref{tab:resSyn}
is that the approach detailed in this thesis generally performs better
than previous approaches across different object types. The largest
improvement in sequence similarity can be observed for the object type
\textit{orders}. This should mean that the object attribute \textit{price}
for object type \textit{orders} is a good feature for the prediction
of suffixes. To prove that assumption the model was trained without the
object attributes, meaning the model is almost identical in strucutre to the
approach by Taymouri \cite{taymouri_encoder-decoder_2020}. As expected, the
results were equal. For the object types items and packages less significant
improvements are observed. Therefore, it is possible the respective object
attributes are less useful features for the model. The absolute results of
the object type \textit{packages} were the best, most likely due to the low
average case length and little variance in possible event flow.

The model is not able to predict suffixes well for the object type
\textit{customer}. Taking into account the statistics from Table
\ref{tab:synStats}, this makes sense. There are only a few cases, making
learning features difficult. Furthermore, the average case length for cases
of object type \textit{orders} is significantly longer than the other object
types' average case length. Very long cases clearly are difficult for the
model to predict. In the case of the synthetic event log, the object type
\textit{customers} is not very useful in predicting an order process, since
customers make multiple orders in the dataset. This can hold interesting
data about the ordering habits of a customer, or can be used to predict
when a customer might pay an order. However, the structure of the model in
this thesis as empirically demonstrated, is better suited to shorter cases,
therefore the metrics of sequence similarity and MAE are very low.

\section{RCLL Dataset}

The RoboCup Logisitcs League (RCLL) is a simulation of factory logistics.
There are two teams of three robots and the goal is to fulfill orders from
a central system. The products that can be ordered consist of the following
components: a base, optional rings, and a cap. There can be between zero
and three rings, but there is always one base and one cap according to the
order. On a whole, the steps are to collect the base, mount rings per the
order, mount the cap, and deliver the order. To achieve this, the robots
work together navigating a map to access all the resources and workstations.
An excerpt of the data can be seen in Figure \ref{fig:rcclFull} in the
Appendix.

\subsection{Description}

The activities modeled in the dataset are:
\begin{center}
    \textit{\{clear-mps, deliver, discard-unknown, fill-cap, fill-rs,
    get-base-to-fill-rs, mount-first-ring, mount-next-ring, process-mps,
    produce-c0, produce-cx\}}
\end{center}

The object types in the RCLL object centric event log:
\begin{center}
    \textit{\{robot, order, products\}}
\end{center}

The attributes for each object type are shown in Table \ref{tab:rcllObjTypes}:
\begin{table}[ht]
    \centering
    \caption{Object attributes in RCLL object centric event log.}
    \label{tab:rcllObjTypes}
    \begin{tabular}{cc} \toprule
    object type & object attribute \\ \midrule
    \textit{robot} & none \\
    \textit{order} & delivery begin, delivery end \\
    \textit{components} & station, cost \\ \bottomrule
    \end{tabular}
\end{table}

The object attributes of delivery begin/end refer to the delivery window
that is associated with the order. In the RCLL simulation early delivery is
considered fatal but late delivery may be acceptable. The object attribute
station and cost only applies if the component is a ring. The station is the
name of a ring station where the particular component needs to be mounted
at and the cost refers to how expensive it is to mount. All object types
are related to all activities in the RCLL data and some statistics are
given in Table \ref{tab:rcllStats}.

\begin{table}[ht]
    \caption{Statistics about RCLL object centric event log.}
    \label{tab:rcllStats}
    \centering
    \begin{tabular}{lllll|lll}\toprule
    &  & \multicolumn{3}{c}{Length of Case} & \multicolumn{3}{c}{Case time (Minutes)} \\
    \cmidrule(lr){3-5}\cmidrule(lr){6-8}
    & Number of cases &  Max & Min & Mean & Max & Min & Mean \\ \midrule
    \textit{robot}      & 3    & 94   & 88   & 91.67  & 16.65  & 4.711 & 11.838 \\ 
    \textit{orders}     & 26   & 16   & 5    &  9.81  & 13.74  & 0.77  & 3.02 \\ 
    \textit{components} & 80   & 18   & 5    & 11.1   & 13.73  & 0.77  & 3.02  \\ \bottomrule
    \end{tabular}
\end{table}

\subsection{Results}

\begin{table}[ht]
    \caption{Sequence similarity and MAE results for RCLL object centric
    event log.}
    \label{tab:resRcll}
    \centering
    \begin{tabular}{lccc|ccc}\toprule
    & \multicolumn{3}{c}{Sequence similariy $S$} & \multicolumn{3}{c}{MAE (normalized)} \\
    \cmidrule(lr){2-4}\cmidrule(lr){5-7}
              & Thesis  & Taymouri \cite{taymouri_encoder-decoder_2020}
              & Tax \cite{tax_predictive_2017} & Thesis &
              Taymouri \cite{taymouri_encoder-decoder_2020} &
              Tax \cite{tax_predictive_2017} \\ \midrule
    \textit{robot}      & 0.1408 & 0.1377   & 0.1128 & 0.4977 & 0.4992 & 0.5182 \\
    \textit{order}      & 0.5242 & 0.4822   & 0.4627 & 0.0325 & 0.0312 & 0.0387 \\
    \textit{components} & 0.4629 & 0.4315   & 0.4179 & 0.0278 & 0.0281 & 0.0391 \\ \bottomrule
    \end{tabular}
\end{table}

The trends observed in the synthetic dataset continue to be present with the
RCLL data, as seen in Table \ref{tab:resRcll}. Overall, in both Sequence
similarity and MAE, the model proposed in this thesis meets or exceeds
the previous research. For the object type \textit{order} the related
object attributes enable the model to achieve a significant improvement
in sequence similarity as opposed to the previous research. The improvement
is less noticeable in the timstamp prediction, which can lead to future
research. The object type \textit{robot} proved to be difficult to predict,
most likely on account of the very long average case lengths. This continues
the trend seen in the synthetic data that the model struggles with longer
cases. Furthermore, since there were no object attributes present for the
object type \textit{robot}, the results are more or less the same as the
approach by Taymouri \cite{taymouri_encoder-decoder_2020}.

\chapter{Conclusion} \label{chap:conclusion}
In this section, the thesis is summarized and an outlook for applications
and expansions are given.

\section{Summary}

In this thesis, using object-centric event logs in state-of-the-art predictive
models was discussed. As demonstrated, the relations that frequently exist
between object types and activities in real processes can not be accurately
modeled by traditional event logs. By not forcing the selection of a single
case id, the OCEL standard enables the preservation of such relations, as well
as storing object attributes. In addition, the object attributes contained
in OCEL data can hold important features for the prediction of next events
and timestamps (suffix) based on an ongoing case (prefix). While this had
been used in past research in the form of case attributes in traditional
event logs, the most recent methods were not utilizing this information.

The model proposed by this thesis augments the recent advancements in
predictive methods with the rich data contained in OCEL. The model makes use of
a Generative Adversarial Network (GAN) in a Sequence to sequence model using
encoder and decoder Long Short-Term Memory layer (LSTM). An individual event
is encoded with activity and timestamp information, as well as the relevant
attributes. Numerical and categorical data is encoded and normalized where
applicable. The model was first trained and evaluated on a synthetic event log
to validate the approach. The synthetic event log represents an order process
with multiple object types, including \textit{customers, items, orders}, and
\textit{packages}. The metrics for testing the model are a measure of sequence
similarity to evaluate the predicted suffix and mean absolute error for the
predicted timestamps. The results showed that overall, the approach presented
in this thesis outperformed the previous approaches in both metrics. For
the object type orders, a noticeable increase in sequence similarity was
achieved. For the object type customers, the results were poor, which can
be explained by the long average case lengths and the low number of cases
available. These trends are also observed on a real event log of RCLL data,
which is a simulation of factory logistics. Again, object types which have
short average case lengths were able to perform better than previous research.

\section{Outlook}

The main goal of the thesis was to use object-centric event logs with
predictive methods. By using object attributes as a feature, the model was
able to outperform past approaches. This can be applied to any OCEL data with
object attributes to improve the prediction methods. However, there are some
drawbacks, the main one being that only activities related to the chosen
object type can be predicted. This can lead to an incomplete or disjointed
view of the underlying process, where intermediate activities that are not
related to an object type can not be predicted by this approach. Therefore,
the main area of future research is developing a method to predict a case
using multiple object types, not just one as in this thesis. This would allow
for all activities to be predicted and can lead to more useful real world
applications. Correlating the disjoint events was the topic of research in
\cite{li_configurable_2018}. By defining correlation profiles, different
views on a process can be gained and such research could be explored in
the future.  However, some more incremental research based on the current
work is possible as well. The encodings chosen for numerical and categorical
values were as simple as possible, to demonstrate the approach working well
independent of a special data preprocessing step. This can be improved by
taking into account business hours as a part of the timestamp prediction
if such data is consistent with those rules. This can lead to significant
increases in the accuracy since the model presented in this thesis could
predict an event to occur ten minutes after closing when it should instead
occur ten minutes after opening the next business day. Furthermore, there
is some recent research to learn the time constraints of events using LSTM
layers in Time2Vec \cite{kazemi_time2vec_2019}.  Instead of determining the
business hours beforehand, Time2Vec can learn the periodic and non-periodic
patterns in the time data. Furthermore, research to determine the most
useful types of object attributes can be conducted. In this thesis, it
was not exhaustively explored why certain attributes could lead to better
results and a more rigorous understanding of the factors that make certain
attributes perform well could lead to insights into how to improve predictions.

\appendix
\chapter{Appendix}
\begin{figure}
    \centering
    \begin{sideways}
    \begin{tabular}{lllllllll} \toprule
    event id & activity & timestamp & items & orders & customers & packages \\ \midrule
    1.0 & place order   & 2019-05-20 09:07:47 & \{880003, 880002, 880001, 880004\} & \{990001\} & \{Marco Pegoraro\} & $\emptyset$ \\
    2.0 & place order   & 2019-05-20 10:35:21 & \{880005, 880006, 880008, 880007\} & \{990002\} & \{Gyunam Park\}    & $\emptyset$ \\
    3.0 & pick item     & 2019-05-20 10:38:17 & \{880006\}                         & \{990002\} & \{Gyunam Park\}    & $\emptyset$ \\
    4.0 & confirm order & 2019-05-20 11:13:54 & \{880003, 880002, 880001, 880004\} & \{990001\} & \{Marco Pegoraro\} & $\emptyset$ \\
    5.0 & pick item     & 2019-05-20 11:20:13 & \{880002\}                         & \{990001\} & \{Marco Pegoraro\} & $\emptyset$ \\
    6.0 & place order   & 2019-05-20 12:30:30 & \{880012, 880011, 880009, 880010\} & \{990003\} & \{Majid Rafiei\}   & $\emptyset$ \\
    7.0 & confirm order & 2019-05-20 12:34:16 & \{880012, 880011, 880009, 880010\} & \{990003\} & \{Majid Rafiei\}   & $\emptyset$ \\ \bottomrule
    \end{tabular}
    \end{sideways}
    \caption{First few events in event table from synthetic OCEL}
    \label{fig:syntheticFull}
\end{figure}

\begin{figure}
    \centering
    \begin{sideways}
    \begin{tabular}{llllll} \toprule
    event id & activity & timestamp & components & order & robot\\ \midrule
    2   & FILL-CAP   & 2020-06-11 15:21:48.150 & \{CAP\_GREY\_1, BASE\_SILVER\_1\} & \{O1\} & \{0\} \\
    6   & PRODUCE-C0 & 2020-06-11 15:22:06.251 & \{CAP\_GREY\_1, BASE\_SILVER\_1\} & \{O1\} & \{2\} \\
    15  & DELIVER    & 2020-06-11 15:22:41.322 & \{CAP\_GREY\_1, BASE\_SILVER\_1\} & \{O1\} & \{2\} \\
    70  & FILL-CAP   & 2020-06-11 15:25:31.418 & \{RING\_YELLOW\_3, CAP\_BLACK\_3, BASE\_SILVER\_3\} & \{O12\} & \{1\} \\
    78  & FILL-RS-1  & 2020-06-11 15:25:57.641 & \{RING\_YELLOW\_3, CAP\_BLACK\_3, BASE\_SILVER\_3\} & \{O12\} &  \{1\} \\
    98  & FILL-CAP   & 2020-06-11 15:27:01.212 & \{BASE\_BLACK\_2, CAP\_GREY\_2\} & \{O11\} & \{0\} \\
    102 & PROCESS-MPS& 2020-06-11 15:27:12.408 & \{BASE\_BLACK\_2, CAP\_GREY\_2\} & \{O11\} & \{0\} \\
    107 & CLEAR-MPS  & 2020-06-11 15:27:26.223 & \{BASE\_BLACK\_2, CAP\_GREY\_2\} & \{O11\} & \{1\} \\
    114 & DELIVER    & 2020-06-11 15:27:48.902 & \{BASE\_BLACK\_2, CAP\_GREY\_2\} & \{O11\} & \{1\} \\
    \end{tabular}
    \end{sideways}
    \caption{First few events in event table from RCLL OCEL}
    \label{fig:rcclFull}
\end{figure}

\bibliographystyle{plain}
\bibliography{references} \addcontentsline{toc}{chapter}{Bibliography}

\begin{thebibliography}{53}
\providecommand{\natexlab}[1]{#1}
\providecommand{\url}[1]{\texttt{#1}}
\expandafter\ifx\csname urlstyle\endcsname\relax
  \providecommand{\doi}[1]{doi: #1}\else
  \providecommand{\doi}{doi: \begingroup \urlstyle{rm}\Url}\fi

\bibitem[van~der Aalst({\natexlab{a}})]{van_der_aalst_process_2016}
Wil van~der Aalst.
\newblock \emph{Process Mining: Data Science in Action}.
\newblock Springer-Verlag, 2 edition, {\natexlab{a}}.
\newblock ISBN 978-3-662-49850-7.
\newblock \doi{10.1007/978-3-662-49851-4}.
\newblock URL \url{https://www.springer.com/gp/book/9783662498507}.

\bibitem[Gilchrist()]{gilchrist_industry_2016}
Alasdair Gilchrist.
\newblock \emph{Industry 4.0}.
\newblock Apress.
\newblock ISBN 978-1-4842-2046-7.
\newblock \doi{10.1007/978-1-4842-2047-4}.

\bibitem[Madakam and Tripathi()]{madakam_internet_2015}
Somayya Madakam and Siddharth Tripathi.
\newblock Internet of things ({IoT}): A literature review.
\newblock 03\penalty0 (5):\penalty0 164.
\newblock \doi{10.4236/jcc.2015.35021}.
\newblock Number: 05 Publisher: Scientific Research Publishing.

\bibitem[Polato et~al.()Polato, Sperduti, Burattin, and
  de~Leoni]{polato_time_2016}
Mirko Polato, Alessandro Sperduti, Andrea Burattin, and Massimiliano de~Leoni.
\newblock Time and activity sequence prediction of business process instances.
\newblock URL \url{http://arxiv.org/abs/1602.07566}.

\bibitem[Mann et~al.(2020)Mann, Pennekamp, Brockhoff, Farhang, Pourbafrani,
  Oster, Uysal, Sharma, Reisgen, Wehrle, et~al.]{mann2020connected}
Samuel Mann, Jan Pennekamp, Tobias Brockhoff, Anahita Farhang, Mahsa
  Pourbafrani, Lukas Oster, Merih~Seran Uysal, Rahul Sharma, Uwe Reisgen, Klaus
  Wehrle, et~al.
\newblock Connected, digitalized welding production—secure, ubiquitous
  utilization of data across process layers.
\newblock In \emph{Advanced Joining Processes}, pages 101--118. Springer, 2020.

\bibitem[Uysal et~al.(2020)Uysal, van Zelst, Brockhoff, Ghahfarokhi,
  Pourbafrani, Schumacher, Junglas, Schuh, and van~der Aalst]{uysal2020process}
Merih~Seran Uysal, Sebastiaan~J van Zelst, Tobias Brockhoff, Anahita~Farhang
  Ghahfarokhi, Mahsa Pourbafrani, Ruben Schumacher, Sebastian Junglas,
  G{\"u}nther Schuh, and WM~van~der Aalst.
\newblock Process mining for production processes in the automotive industry.
\newblock In \emph{Industry Forum at BPM}, volume~20, 2020.

\bibitem[Nik et~al.(2019)Nik, van~der Aalst, and Sani]{nik2019bipm}
Mohammad Reza~Harati Nik, Wil~MP van~der Aalst, and Mohammadreza~Fani Sani.
\newblock Bipm: Combining bi and process mining.
\newblock In \emph{DATA}, pages 123--128, 2019.

\bibitem[Evermann et~al.()Evermann, Rehse, and
  Fettke]{evermann_predicting_2017}
Joerg Evermann, Jana-Rebecca Rehse, and Peter Fettke.
\newblock Predicting process behaviour using deep learning.
\newblock 100:\penalty0 129--140.
\newblock ISSN 01679236.
\newblock \doi{10.1016/j.dss.2017.04.003}.
\newblock URL \url{http://arxiv.org/abs/1612.04600}.

\bibitem[Folino et~al.()Folino, Guarascio, and
  Pontieri]{folino_discovering_2012}
Francesco Folino, Massimo Guarascio, and Luigi Pontieri.
\newblock Discovering context-aware models for predicting business process
  performances.
\newblock In \emph{On the Move to Meaningful Internet Systems: {OTM} 2012},
  Lecture Notes in Computer Science, pages 287--304. Springer.
\newblock ISBN 978-3-642-33606-5.
\newblock \doi{10.1007/978-3-642-33606-5_18}.

\bibitem[Breuker et~al.()Breuker, Matzner, Delfmann, and
  Becker]{breuker_comprehensible_2016}
Dominic Breuker, Martin Matzner, Patrick Delfmann, and Jörg Becker.
\newblock Comprehensible predictive models for business processes.
\newblock 40\penalty0 (4):\penalty0 1009--1034.
\newblock ISSN 0276-7783.
\newblock \doi{10.25300/MISQ/2016/40.4.10}.

\bibitem[noa()]{noauthor_ieee_2016}
{IEEE} standard for {eXtensible} event stream ({XES}) for achieving
  interoperability in event logs and event streams.
\newblock pages 1--50.
\newblock \doi{10.1109/IEEESTD.2016.7740858}.
\newblock Conference Name: {IEEE} Std 1849-2016.

\bibitem[Pasquadibisceglie et~al.()Pasquadibisceglie, Appice, Castellano, and
  Malerba]{pasquadibisceglie_using_2019}
Vincenzo Pasquadibisceglie, Annalisa Appice, Giovanna Castellano, and Donato
  Malerba.
\newblock Using convolutional neural networks for predictive process analytics.
\newblock In \emph{2019 International Conference on Process Mining ({ICPM})},
  pages 129--136.
\newblock \doi{10.1109/ICPM.2019.00028}.

\bibitem[Tax et~al.()Tax, Verenich, La~Rosa, and Dumas]{tax_predictive_2017}
Niek Tax, Ilya Verenich, Marcello La~Rosa, and Marlon Dumas.
\newblock Predictive business process monitoring with {LSTM} neural networks.
\newblock 10253:\penalty0 477--492.
\newblock \doi{10.1007/978-3-319-59536-8_30}.
\newblock URL \url{http://arxiv.org/abs/1612.02130}.

\bibitem[Taymouri et~al.()Taymouri, La~Rosa, Erfani, Bozorgi, and
  Verenich]{taymouri_predictive_2020}
Farbod Taymouri, Marcello La~Rosa, Sarah Erfani, Zahra~Dasht Bozorgi, and Ilya
  Verenich.
\newblock Predictive business process monitoring via generative adversarial
  nets: The case of next event prediction.
\newblock URL \url{http://arxiv.org/abs/2003.11268}.

\bibitem[Taymouri and La~Rosa()]{taymouri_encoder-decoder_2020}
Farbod Taymouri and Marcello La~Rosa.
\newblock Encoder-decoder generative adversarial nets for suffix generation and
  remaining time prediction of business process models.
\newblock URL \url{http://arxiv.org/abs/2007.16030}.

\bibitem[van~der Aalst({\natexlab{b}})]{van_der_aalst_object-centric_2019}
Wil van~der Aalst.
\newblock Object-centric process mining: Dealing with divergence and
  convergence in event data.
\newblock In \emph{Software Engineering and Formal Methods}, Lecture Notes in
  Computer Science, pages 3--25. Springer International Publishing,
  {\natexlab{b}}.
\newblock ISBN 978-3-030-30446-1.
\newblock \doi{10.1007/978-3-030-30446-1_1}.

\bibitem[Ghahfarokhi et~al.(2021{\natexlab{a}})Ghahfarokhi, Park, Berti, and
  van~der Aalst]{ghahfarokhi2021ocel}
Anahita~Farhang Ghahfarokhi, Gyunam Park, Alessandro Berti, and Wil~MP van~der
  Aalst.
\newblock Ocel: A standard for object-centric event logs.
\newblock In \emph{European Conference on Advances in Databases and Information
  Systems}, pages 169--175. Springer, 2021{\natexlab{a}}.

\bibitem[Ghahfarokhi et~al.(2021{\natexlab{b}})Ghahfarokhi, Berti, and van~der
  Aalst]{ghahfarokhi2021process}
Anahita~Farhang Ghahfarokhi, Alessandro Berti, and Wil~MP van~der Aalst.
\newblock Process comparison using object-centric process cubes.
\newblock \emph{arXiv preprint arXiv:2103.07184}, 2021{\natexlab{b}}.

\bibitem[Farhang~Ghahfarokhi and van~der Aalst(2022)]{farhang2022python}
Anahita Farhang~Ghahfarokhi and Wil~MP van~der Aalst.
\newblock A python tool for object-centric process mining comparison.
\newblock \emph{arXiv e-prints}, pages arXiv--2202, 2022.

\bibitem[Berti et~al.(2022)Berti, Ghahfarokhi, Park, and van~der
  Aalst]{berti2022scalable}
Alessandro Berti, Anahita~Farhang Ghahfarokhi, Gyunam Park, and Wil~MP van~der
  Aalst.
\newblock A scalable database for the storage of object-centric event logs.
\newblock \emph{arXiv preprint arXiv:2202.05639}, 2022.

\bibitem[Li et~al.({\natexlab{a}})Li, de~Murillas, de~Carvalho, and van~der
  Aalst]{li_extracting_2018}
Guangming Li, Eduardo González~López de~Murillas, Renata~Medeiros
  de~Carvalho, and Wil M.~P. van~der Aalst.
\newblock Extracting object-centric event logs to support process mining on
  databases.
\newblock In \emph{Information Systems in the Big Data Era}, Lecture Notes in
  Business Information Processing, pages 182--199. Springer International
  Publishing, {\natexlab{a}}.
\newblock ISBN 978-3-319-92901-9.
\newblock \doi{10.1007/978-3-319-92901-9_16}.

\bibitem[van~der Aalst et~al.({\natexlab{a}})van~der Aalst, Schonenberg, and
  Song]{van_der_aalst_time_2011}
Wil van~der Aalst, M.~H. Schonenberg, and M.~Song.
\newblock Time prediction based on process mining.
\newblock 36\penalty0 (2):\penalty0 450--475, {\natexlab{a}}.
\newblock ISSN 0306-4379.
\newblock \doi{10.1016/j.is.2010.09.001}.
\newblock URL
  \url{http://www.sciencedirect.com/science/article/pii/S0306437910000864}.

\bibitem[Ghahfarokhi et~al.()Ghahfarokhi, Park, Berti, and van~der
  Aalst]{ghahfarokhi_ocel_2020}
Anahita~Farhang Ghahfarokhi, Gyunam Park, Alessandro Berti, and Wil van~der
  Aalst.
\newblock {OCEL} standard.
\newblock URL \url{http://ocel-standard.org/1.0/specification.pdf}.

\bibitem[van Dongen et~al.()van Dongen, Crooy, and van~der
  Aalst]{van_dongen_cycle_2008}
B.~F. van Dongen, R.~A. Crooy, and Wil van~der Aalst.
\newblock Cycle time prediction: When will this case finally be finished?
\newblock In \emph{On the Move to Meaningful Internet Systems: {OTM} 2008},
  Lecture Notes in Computer Science, pages 319--336. Springer.
\newblock ISBN 978-3-540-88871-0.
\newblock \doi{10.1007/978-3-540-88871-0_22}.

\bibitem[Senderovich et~al.()Senderovich, Di~Francescomarino, Ghidini, Jorbina,
  and Maggi]{senderovich_intra_2017}
Arik Senderovich, Chiara Di~Francescomarino, Chiara Ghidini, Kerwin Jorbina,
  and Fabrizio~Maria Maggi.
\newblock Intra and inter-case features in predictive process monitoring: A
  tale of two dimensions.
\newblock In \emph{Business Process Management}, Lecture Notes in Computer
  Science, pages 306--323. Springer International Publishing.
\newblock ISBN 978-3-319-65000-5.
\newblock \doi{10.1007/978-3-319-65000-5_18}.

\bibitem[Schmidhuber()]{schmidhuber_deep_2015}
Juergen Schmidhuber.
\newblock Deep learning in neural networks.
\newblock 61:\penalty0 85--117.
\newblock ISSN 08936080.
\newblock \doi{10.1016/j.neunet.2014.09.003}.
\newblock URL \url{http://arxiv.org/abs/1404.7828}.

\bibitem[Nielsen()]{nielsen_neural_2015}
Michael Nielsen.
\newblock \emph{Neural networks and deep learning}.
\newblock Determination Press.

\bibitem[Yan()]{yan_understanding_2017}
Shi Yan.
\newblock Understanding {LSTM} and its diagrams.
\newblock URL
  \url{https://medium.com/mlreview/understanding-lstm-and-its-diagrams-37e2f46f1714}.

\bibitem[Hochreiter and Schmidhuber()]{hochreiter_long_1997}
Sepp Hochreiter and Jürgen Schmidhuber.
\newblock Long short-term memory.
\newblock 9\penalty0 (8):\penalty0 1735--1780.
\newblock ISSN 0899-7667.
\newblock \doi{10.1162/neco.1997.9.8.1735}.
\newblock Publisher: {MIT} Press.

\bibitem[Goodfellow et~al.()Goodfellow, Pouget-Abadie, Mirza, Xu, and
  Warde-Farley]{goodfellow_generative_2014}
Ian Goodfellow, Jean Pouget-Abadie, Mehdi Mirza, Bing Xu, and David
  Warde-Farley.
\newblock Generative adversarial nets.
\newblock In \emph{Advances in Neural Information Processing Systems 27}, pages
  2672--2680. Curran Associates, Inc.

\bibitem[Radford et~al.()Radford, Metz, and
  Chintala]{radford_unsupervised_2016}
Alec Radford, Luke Metz, and Soumith Chintala.
\newblock Unsupervised representation learning with deep convolutional
  generative adversarial networks.
\newblock URL \url{http://arxiv.org/abs/1511.06434}.

\bibitem[Sutskever et~al.()Sutskever, Vinyals, and Le]{sutskever_sequence_2014}
Ilya Sutskever, Oriol Vinyals, and Quoc~V. Le.
\newblock Sequence to sequence learning with neural networks.
\newblock URL \url{http://arxiv.org/abs/1409.3215}.

\bibitem[Agrawal et~al.()Agrawal, Gunopulos, and Leymann]{agrawal_mining_1998}
Rakesh Agrawal, Dimitrios Gunopulos, and Frank Leymann.
\newblock Mining process models from workflow logs.
\newblock In \emph{Advances in Database Technology — {EDBT}'98}, Lecture
  Notes in Computer Science, pages 467--483. Springer.
\newblock ISBN 978-3-540-69709-1.
\newblock \doi{10.1007/BFb0101003}.

\bibitem[van~der Aalst et~al.({\natexlab{b}})van~der Aalst, Weijters, and
  Maruster]{van_der_aalst_workflow_2004}
Wil van~der Aalst, Ton Weijters, and Laura Maruster.
\newblock Workflow mining: Discovering process models from event logs.
\newblock 16\penalty0 (9):\penalty0 1128--1142, {\natexlab{b}}.
\newblock \doi{10.1109/TKDE.2004.47}.

\bibitem[Weijters and van~der Aalst()]{weijters_rediscovering_2003}
A.~Weijters and Wil van~der Aalst.
\newblock Rediscovering workflow models from event-based data using little
  thumb.
\newblock 10:\penalty0 151--162.
\newblock \doi{10.3233/ICA-2003-10205}.

\bibitem[Le et~al.()Le, Gabrys, and Nauck]{le_hybrid_2012}
Mai Le, Bogdan Gabrys, and Detlef Nauck.
\newblock A hybrid model for business process event prediction.
\newblock In \emph{Research and Development in Intelligent Systems {XXIX}},
  pages 179--192. Springer.
\newblock ISBN 978-1-4471-4739-8.
\newblock \doi{10.1007/978-1-4471-4739-8_13}.

\bibitem[Lakshmanan et~al.()Lakshmanan, Shamsi, Doganata, Unuvar, and
  Khalaf]{lakshmanan_markov_2015}
Geetika~T. Lakshmanan, Davood Shamsi, Yurdaer~N. Doganata, Merve Unuvar, and
  Rania Khalaf.
\newblock A markov prediction model for data-driven semi-structured business
  processes.
\newblock 42\penalty0 (1):\penalty0 97--126.
\newblock ISSN 0219-3116.
\newblock \doi{10.1007/s10115-013-0697-8}.

\bibitem[Rogge-Solti and Weske()]{rogge-solti_prediction_2013}
Andreas Rogge-Solti and Mathias Weske.
\newblock Prediction of remaining service execution time using stochastic petri
  nets with arbitrary firing delays.
\newblock In \emph{Service-Oriented Computing}, Lecture Notes in Computer
  Science, pages 389--403. Springer.
\newblock ISBN 978-3-642-45005-1.
\newblock \doi{10.1007/978-3-642-45005-1_27}.

\bibitem[Verenich et~al.()Verenich, Dumas, Rosa, Maggi, and
  Teinemaa]{verenich_survey_2019}
Ilya Verenich, Marlon Dumas, Marcello~La Rosa, Fabrizio~Maria Maggi, and Irene
  Teinemaa.
\newblock Survey and cross-benchmark comparison of remaining time prediction
  methods in business process monitoring.
\newblock 10\penalty0 (4):\penalty0 34:1--34:34.
\newblock ISSN 2157-6904.
\newblock \doi{10.1145/3331449}.

\bibitem[Camargo et~al.()Camargo, Dumas, and
  González-Rojas]{camargo_learning_2019}
Manuel Camargo, Marlon Dumas, and Oscar González-Rojas.
\newblock Learning accurate {LSTM} models of business processes.
\newblock In \emph{Business Process Management}, Lecture Notes in Computer
  Science, pages 286--302. Springer International Publishing.
\newblock ISBN 978-3-030-26619-6.
\newblock \doi{10.1007/978-3-030-26619-6_19}.

\bibitem[Berti and van~der Aalst()]{berti_extracting_2020}
Alessandro Berti and Wil van~der Aalst.
\newblock Extracting multiple viewpoint models from relational databases.
\newblock URL \url{http://arxiv.org/abs/2001.02562}.

\bibitem[Simović et~al.()Simović, Babarogić, and
  Pantelić]{simovic_domain-specific_2018}
A.~P. Simović, S.~Babarogić, and O.~Pantelić.
\newblock A domain-specific language for supporting event log extraction from
  {ERP} systems.
\newblock In \emph{2018 7th International Conference on Computers
  Communications and Control ({ICCCC})}, pages 12--16.
\newblock \doi{10.1109/ICCCC.2018.8390430}.

\bibitem[de~Murillas et~al.()de~Murillas, Reijers, and van~der
  Aalst]{de_murillas_case_2020}
E.~González~López de~Murillas, H.~A. Reijers, and W.~M.~P. van~der Aalst.
\newblock Case notion discovery and recommendation: automated event log
  building on databases.
\newblock 62\penalty0 (7):\penalty0 2539--2575.
\newblock ISSN 0219-3116.
\newblock \doi{10.1007/s10115-019-01430-6}.

\bibitem[van~der Aalst and Berti()]{van_der_aalst_discovering_2020}
Wil M.~P. van~der Aalst and Alessandro Berti.
\newblock Discovering object-centric petri nets.
\newblock URL \url{http://arxiv.org/abs/2010.02047}.

\bibitem[Cohn and Hull()]{cohn_business_2009}
David Cohn and Richard Hull.
\newblock Business artifacts: A data-centric approach to modeling business
  operations and processes.
\newblock 32:\penalty0 3--9.

\bibitem[Narendra et~al.()Narendra, Badr, Thiran, and
  Maamar]{narendra_towards_2009}
N.~C. Narendra, Y.~Badr, P.~Thiran, and Z.~Maamar.
\newblock Towards a unified approach for business process modeling using
  context-based artifacts and web services.
\newblock In \emph{2009 {IEEE} International Conference on Services Computing},
  pages 332--339.
\newblock \doi{10.1109/SCC.2009.14}.

\bibitem[Li et~al.({\natexlab{b}})Li, de~Carvalho, and van~der
  Aalst]{li_object-centric_2019}
Guangming Li, Renata~Medeiros de~Carvalho, and Wil M.~P. van~der Aalst.
\newblock Object-centric behavioral constraint models: a hybrid model for
  behavioral and data perspectives.
\newblock In \emph{Proceedings of the 34th {ACM}/{SIGAPP} Symposium on Applied
  Computing}, {SAC} '19, pages 48--56. Association for Computing Machinery,
  {\natexlab{b}}.
\newblock ISBN 978-1-4503-5933-7.
\newblock \doi{10.1145/3297280.3297287}.

\bibitem[Jang et~al.()Jang, Gu, and Poole]{jang_categorical_2017}
Eric Jang, Shixiang Gu, and Ben Poole.
\newblock Categorical reparameterization with gumbel-softmax.
\newblock URL \url{http://arxiv.org/abs/1611.01144}.

\bibitem[Maddison et~al.()Maddison, Tarlow, and Minka]{maddison__2015}
Chris~J. Maddison, Daniel Tarlow, and Tom Minka.
\newblock A* sampling.
\newblock URL \url{http://arxiv.org/abs/1411.0030}.

\bibitem[Sønderby et~al.()Sønderby, Caballero, Theis, Shi, and
  Huszár]{sonderby_amortised_2017}
Casper~Kaae Sønderby, Jose Caballero, Lucas Theis, Wenzhe Shi, and Ferenc
  Huszár.
\newblock Amortised {MAP} inference for image super-resolution.
\newblock URL \url{http://arxiv.org/abs/1610.04490}.

\bibitem[Williams and Zipser()]{williams_learning_1989}
Ronald~J. Williams and David Zipser.
\newblock A learning algorithm for continually running fully recurrent neural
  networks.
\newblock 1\penalty0 (2):\penalty0 270--280.
\newblock ISSN 0899-7667.
\newblock \doi{10.1162/neco.1989.1.2.270}.

\bibitem[Li et~al.({\natexlab{c}})Li, M.~de Carvalho, and
  Aalst]{li_configurable_2018}
Guangming Li, Renata M.~de Carvalho, and Wil Aalst.
\newblock Configurable event correlation for process discovery from
  object-centric event data.
\newblock pages 203--210, {\natexlab{c}}.
\newblock \doi{10.1109/ICWS.2018.00033}.

\bibitem[Kazemi et~al.()Kazemi, Goel, Eghbali, Ramanan, Sahota, Thakur, Wu,
  Smyth, Poupart, and Brubaker]{kazemi_time2vec_2019}
Seyed~Mehran Kazemi, Rishab Goel, Sepehr Eghbali, Janahan Ramanan, Jaspreet
  Sahota, Sanjay Thakur, Stella Wu, Cathal Smyth, Pascal Poupart, and Marcus
  Brubaker.
\newblock Time2vec: Learning a vector representation of time.
\newblock URL \url{http://arxiv.org/abs/1907.05321}.

\end{thebibliography}

\end{document}